\documentclass[runningheads]{llncs}
\usepackage[T1]{fontenc}
\usepackage{graphicx}
\usepackage{booktabs}
\usepackage{multirow,array}
\usepackage{amsmath,amssymb}
\usepackage{algorithm}
\usepackage{algorithmic}
\usepackage{xcolor}
\usepackage{tikz}
\usetikzlibrary{arrows.meta,positioning}
\usepackage{url}
\usepackage{verbatim}
\usepackage{placeins}
\usepackage{listings}
\usepackage[hidelinks]{hyperref}
\hypersetup{pdftitle={Explanation-Guided Evolution of LLM-driven Feature Engineering},pdfauthor={Jiahuizi Luo, Thomas Bäck, Niki van Stein}}

\newcommand{\shapevo}{SHAPEvo-FE}
\newcommand{\llmfe}{LLM-FE}

\begin{document}

\title{Explanation-Guided Evolution of LLM-driven Feature Engineering\thanks{Preprint of a paper under review. The supplementary material submitted with the paper is included as an appendix.}}
\titlerunning{Explanation-Guided Evolution of LLM-driven Feature Engineering}
\author{Jiahuizi Luo\orcidID{0009-0007-4341-0164} \and Thomas B\"ack \and Niki van~Stein\orcidID{0000-0002-0013-7969}}
\authorrunning{J. Luo et al.}
\institute{LIACS, Leiden University, Leiden, The Netherlands\\
\email{luoj5@vuw.leidenuniv.nl, T.H.W.Baeck@liacs.leidenuniv.nl, n.van.stein@liacs.leidenuniv.nl}}
\maketitle

\begin{abstract}
Large language models are increasingly used as evolutionary variation operators to write feature-engineering programs for tabular data. The state-of-the-art method, \llmfe, shows the LLM only the code and score of earlier programs, so the search never learns which of the generated features the downstream model actually uses. We present \shapevo, a $(\mu{+}\lambda)$ evolution strategy in which TreeSHAP attributions of the fitted model are turned into a short feedback report that labels each generated feature ``keep'', ``refine'' or ``replace'', which is used in the next mutation prompt. We compare the two methods under a controlled protocol, with the same LLM backbone, budget, cross-validation splits, completion limit and evaluation code, on 22 datasets and three different backbones. With a capable backbone, SHAP guidance makes the search significantly cheaper. With DeepSeek-V4-Flash and Claude-Haiku-5.5, \shapevo{} reaches \llmfe's final validation level with 38 and 62 programs on the median dataset, against about 180 for \llmfe, in a third to a half of the wall-clock time and with about a third of the LLM tokens, and its normalised area under the best-so-far curve is larger on 20 and 17 of 22 datasets respectively (Wilcoxon $p<0.001$ (DeepSeek) and $p=0.003$ (Haiku)). The same loop without the SHAP report needs about twice as many programs. With GPT-3.5, which ignores the report, \llmfe{} searches faster. SHAP feedback helps when the backbone follows it and makes no difference when the backbone ignores it.

\end{abstract}

%======================================================================
\section{Introduction}
\label{sec:intro}

Feature engineering still determines much of the accuracy of tabular models~\cite{domingos2012useful}. Gradient-boosted trees on well-designed features rival or beat deep models~\cite{grinsztajn2022tree}. Several recent methods let a large language model (LLM) write the Python programs that transform the raw columns and wrap this generation in an evolutionary loop~\cite{hollmann2024caafe,nam2024octree,abhyankar2025llmfe}. The strongest of them, \llmfe~\cite{abhyankar2025llmfe}, evolves such programs with an island-model experience buffer in the style of FunSearch~\cite{romeraparedes2024funsearch} and reports the best results on a broad benchmark of classification and regression datasets.

In \llmfe{} the LLM sees the code and the validation score of a few earlier programs. It does not see which of the generated features the downstream model actually uses, although the fitted model already holds this information and TreeSHAP can extract it at a small extra cost. A program that adds five features of which one matters looks, in the prompt, exactly like a program whose five features all contribute. We hypothesize that feeding model explanations back into the prompt can let the LLM edit more deliberately, and therefore search more effectively. It could keep the features that work, rewrite the ones that do not, and build on inputs the model already treats as interacting.

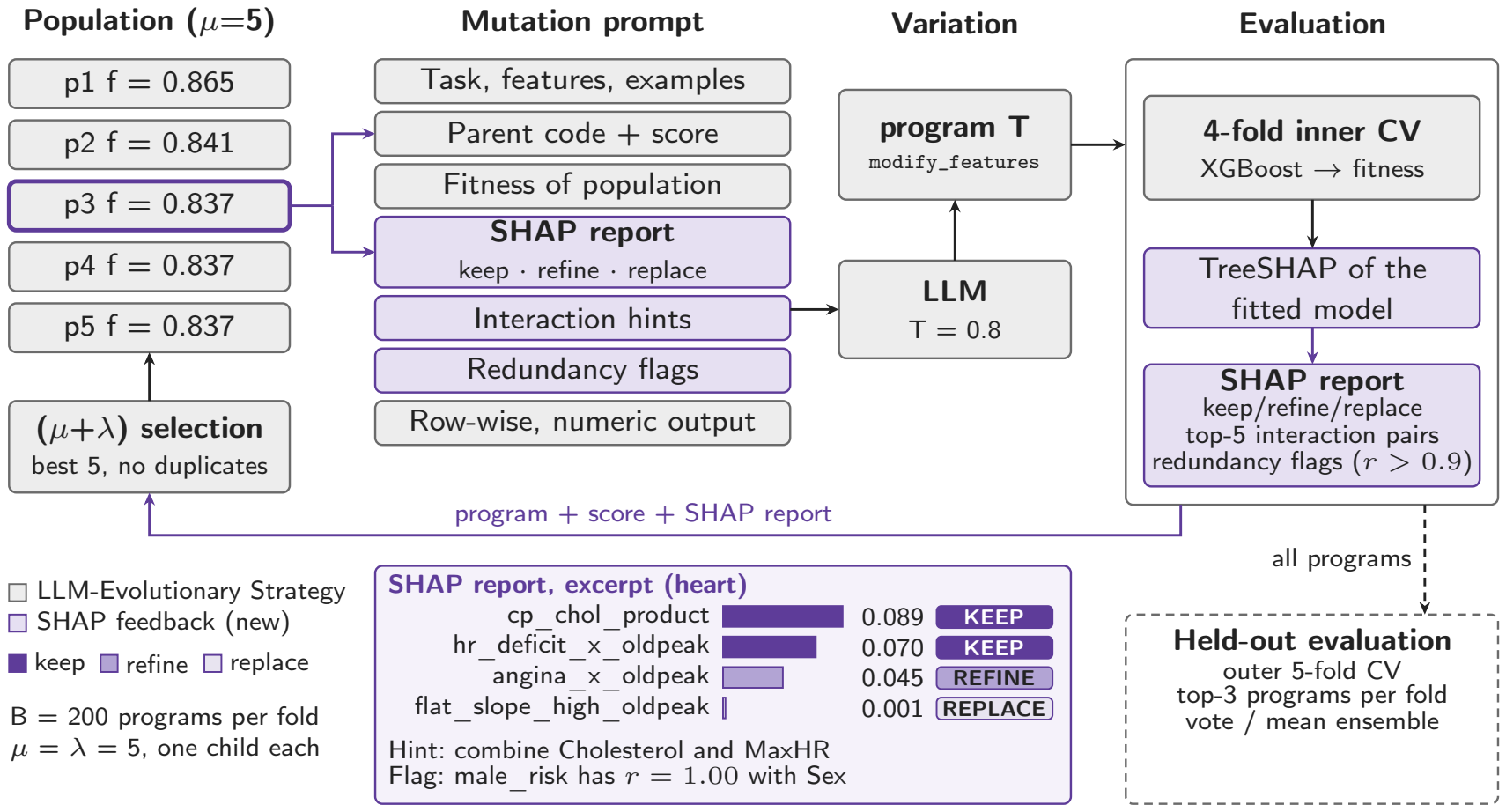
\begin{figure}[t]
\centering
\resizebox{\textwidth}{!}{% Fig. 1 of SHAPEvo-FE. Needs: \usepackage{tikz} \usetikzlibrary{arrows.meta}
% Usage: \begin{figure}\centering\input{fig1_shapevo_tikz}\caption{...}\end{figure}
% Drawn for a width of 12.2 cm (LNCS text width); wrap in \resizebox{\textwidth}{!}{...} if needed.
\definecolor{sfgrey}{HTML}{EDEDED}
\definecolor{sfgline}{HTML}{6B6B6B}
\definecolor{sfpur}{HTML}{5E3C99}
\definecolor{sfpurfill}{HTML}{E6E0F2}
\definecolor{sfpurlt}{HTML}{F5F2FA}
\definecolor{sfmid}{HTML}{B2A4D4}
\definecolor{sflt}{HTML}{E8E4F0}
\definecolor{sfink}{HTML}{222222}
\definecolor{sfwhite}{HTML}{FFFFFF}
\begin{tikzpicture}[x=1cm,y=1cm,
  b55/.style={font=\sffamily\bfseries\fontsize{5.5}{6.3}\selectfont, inner sep=0pt},
  b65/.style={font=\sffamily\bfseries\fontsize{6.5}{7.5}\selectfont, inner sep=0pt},
  b7/.style={font=\sffamily\bfseries\fontsize{7}{8.0}\selectfont, inner sep=0pt},
  m55/.style={font=\ttfamily\fontsize{5.5}{6.3}\selectfont, inner sep=0pt},
  s6/.style={font=\sffamily\fontsize{6}{6.9}\selectfont, inner sep=0pt},
  s65/.style={font=\sffamily\fontsize{6.5}{7.5}\selectfont, inner sep=0pt},
  s7/.style={font=\sffamily\fontsize{7}{8.0}\selectfont, inner sep=0pt}]
\node[b7, text=sfink] at (1.15,6.45) {Population ($\mu$=5)};
\node[b7, text=sfink] at (4.70,6.45) {Mutation prompt};
\node[b7, text=sfink] at (7.75,6.45) {Variation};
\node[b7, text=sfink] at (10.68,6.45) {Evaluation};
\draw[fill=sfgrey, draw=sfgline, line width=0.5pt, rounded corners=1.7pt] (0.00,5.75) rectangle (2.30,6.15);
\node[s7, text=sfink] at (1.15,5.95) {p1   f = 0.865};
\draw[fill=sfgrey, draw=sfgline, line width=0.5pt, rounded corners=1.7pt] (0.00,5.25) rectangle (2.30,5.65);
\node[s7, text=sfink] at (1.15,5.45) {p2   f = 0.841};
\draw[fill=sfgrey, draw=sfpur, line width=1.0pt, rounded corners=1.7pt] (0.00,4.75) rectangle (2.30,5.15);
\node[s7, text=sfink] at (1.15,4.95) {p3   f = 0.837};
\draw[fill=sfgrey, draw=sfgline, line width=0.5pt, rounded corners=1.7pt] (0.00,4.25) rectangle (2.30,4.65);
\node[s7, text=sfink] at (1.15,4.45) {p4   f = 0.837};
\draw[fill=sfgrey, draw=sfgline, line width=0.5pt, rounded corners=1.7pt] (0.00,3.75) rectangle (2.30,4.15);
\node[s7, text=sfink] at (1.15,3.95) {p5   f = 0.837};
\draw[fill=sfgrey, draw=sfgline, line width=0.5pt, rounded corners=1.7pt] (0.00,2.60) rectangle (2.30,3.35);
\node[b7, text=sfink] at (1.15,3.11) {($\mu$+$\lambda$) selection};
\node[s6, text=sfink] at (1.15,2.81) {best 5, no duplicates};
\draw[-{Stealth[length=3pt,width=2.6pt]}, draw=sfink, line width=0.6pt] (1.15,3.35) -- (1.15,3.75);
\draw[fill=sfgrey, draw=sfgline, line width=0.5pt, rounded corners=1.7pt] (3.00,5.79) rectangle (6.40,6.15);
\node[s7, text=sfink] at (4.70,5.97) {Task, features, examples};
\draw[fill=sfgrey, draw=sfgline, line width=0.5pt, rounded corners=1.7pt] (3.00,5.36) rectangle (6.40,5.72);
\node[s7, text=sfink] at (4.70,5.54) {Parent code + score};
\draw[fill=sfgrey, draw=sfgline, line width=0.5pt, rounded corners=1.7pt] (3.00,4.93) rectangle (6.40,5.29);
\node[s7, text=sfink] at (4.70,5.11) {Fitness of population};
\draw[fill=sfpurfill, draw=sfpur, line width=0.5pt, rounded corners=1.7pt] (3.00,4.28) rectangle (6.40,4.86);
\node[b7, text=sfink] at (4.70,4.72) {SHAP report};
\node[s65, text=sfink] at (4.70,4.41) {keep $\cdot$ refine $\cdot$ replace};
\draw[fill=sfpurfill, draw=sfpur, line width=0.5pt, rounded corners=1.7pt] (3.00,3.85) rectangle (6.40,4.21);
\node[s7, text=sfink] at (4.70,4.03) {Interaction hints};
\draw[fill=sfpurfill, draw=sfpur, line width=0.5pt, rounded corners=1.7pt] (3.00,3.42) rectangle (6.40,3.78);
\node[s7, text=sfink] at (4.70,3.60) {Redundancy flags};
\draw[fill=sfgrey, draw=sfgline, line width=0.5pt, rounded corners=1.7pt] (3.00,2.99) rectangle (6.40,3.35);
\node[s7, text=sfink] at (4.70,3.17) {Row-wise, numeric output};
\draw[draw=sfpur, line width=0.6pt] (2.30,4.95) -- (2.65,4.95);
\draw[-{Stealth[length=3pt,width=2.6pt]}, draw=sfpur, line width=0.6pt] (2.65,4.95) -- (2.65,5.54) -- (3.00,5.54);
\draw[-{Stealth[length=3pt,width=2.6pt]}, draw=sfpur, line width=0.6pt] (2.65,4.95) -- (2.65,4.57) -- (3.00,4.57);
\draw[fill=sfgrey, draw=sfgline, line width=0.5pt, rounded corners=1.7pt] (6.80,3.70) rectangle (8.70,4.50);
\node[b7, text=sfink] at (7.75,4.25) {LLM};
\node[s65, text=sfink] at (7.75,3.94) {T = 0.8};
\draw[fill=sfgrey, draw=sfgline, line width=0.5pt, rounded corners=1.7pt] (6.80,5.00) rectangle (8.70,5.90);
\node[b7, text=sfink] at (7.75,5.58) {program T};
\node[m55, text=sfink] at (7.75,5.29) {modify\_features};
\draw[-{Stealth[length=3pt,width=2.6pt]}, draw=sfink, line width=0.6pt] (6.40,4.10) -- (6.80,4.10);
\draw[-{Stealth[length=3pt,width=2.6pt]}, draw=sfink, line width=0.6pt] (7.75,4.50) -- (7.75,5.00);
\draw[-{Stealth[length=3pt,width=2.6pt]}, draw=sfink, line width=0.6pt] (8.70,5.45) -- (9.15,5.45);
\draw[fill=sfwhite, draw=sfgline, line width=0.5pt, rounded corners=1.7pt] (9.15,2.50) rectangle (12.20,6.15);
\draw[fill=sfgrey, draw=sfgline, line width=0.5pt, rounded corners=1.7pt] (9.30,5.00) rectangle (12.05,5.85);
\node[b7, text=sfink] at (10.68,5.57) {4-fold inner CV};
\node[s65, text=sfink] at (10.68,5.26) {XGBoost $\rightarrow$ fitness};
\draw[fill=sfpurfill, draw=sfpur, line width=0.5pt, rounded corners=1.7pt] (9.30,3.95) rectangle (12.05,4.60);
\node[s7, text=sfink] at (10.68,4.44) {TreeSHAP of the};
\node[s7, text=sfink] at (10.68,4.11) {fitted model};
\draw[fill=sfpurfill, draw=sfpur, line width=0.5pt, rounded corners=1.7pt] (9.30,2.65) rectangle (12.05,3.65);
\node[b7, text=sfink] at (10.68,3.50) {SHAP report};
\node[s6, text=sfink] at (10.68,3.28) {keep/refine/replace};
\node[s6, text=sfink] at (10.68,3.06) {top-5 interaction pairs};
\node[s6, text=sfink] at (10.68,2.84) {redundancy flags ($r>0.9$)};
\draw[-{Stealth[length=3pt,width=2.6pt]}, draw=sfink, line width=0.6pt] (10.68,5.00) -- (10.68,4.60);
\draw[-{Stealth[length=3pt,width=2.6pt]}, draw=sfpur, line width=0.6pt] (10.68,3.95) -- (10.68,3.65);
\draw[-{Stealth[length=3pt,width=2.6pt]}, draw=sfpur, line width=0.6pt] (9.60,2.50) -- (9.60,2.25) -- (1.15,2.25) -- (1.15,2.60);
\node[s6, text=sfpur] at (5.20,2.42) {program + score + SHAP report};
\draw[fill=sfwhite, draw=sfgline, line width=0.5pt, dash pattern=on 2pt off 1.4pt, rounded corners=1.7pt] (9.15,0.05) rectangle (12.20,1.60);
\node[b7, text=sfink] at (10.68,1.40) {Held-out evaluation};
\node[s6, text=sfink] at (10.68,1.15) {outer 5-fold CV};
\node[s6, text=sfink] at (10.68,0.93) {top-3 programs per fold};
\node[s6, text=sfink] at (10.68,0.71) {vote / mean ensemble};
\draw[-{Stealth[length=3pt,width=2.6pt]}, draw=sfink, line width=0.6pt, dash pattern=on 2pt off 1.4pt] (11.60,2.50) -- (11.60,1.60);
\node[s6, text=sfink, anchor=east] at (11.50,2.05) {all programs};
\draw[fill=sfpurlt, draw=sfpur, line width=0.5pt, rounded corners=1.7pt] (3.00,0.05) rectangle (8.70,2.00);
\node[b65, text=sfpur, anchor=west] at (3.10,1.83) {SHAP report, excerpt (heart)};
\node[s6, text=sfink, anchor=east] at (5.75,1.58) {cp\_chol\_product};
\draw[fill=sfpur, draw=sfpur, line width=0.4pt, rounded corners=0.0pt] (5.85,1.50) rectangle (6.83,1.67);
\node[s6, text=sfink, anchor=west] at (6.98,1.58) {0.089};
\draw[fill=sfpur, draw=sfpur, line width=0.4pt, rounded corners=1.1pt] (7.60,1.49) rectangle (8.55,1.67);
\node[b55, text=sfwhite] at (8.07,1.58) {KEEP};
\node[s6, text=sfink, anchor=east] at (5.75,1.33) {hr\_deficit\_x\_oldpeak};
\draw[fill=sfpur, draw=sfpur, line width=0.4pt, rounded corners=0.0pt] (5.85,1.25) rectangle (6.61,1.42);
\node[s6, text=sfink, anchor=west] at (6.98,1.33) {0.070};
\draw[fill=sfpur, draw=sfpur, line width=0.4pt, rounded corners=1.1pt] (7.60,1.24) rectangle (8.55,1.42);
\node[b55, text=sfwhite] at (8.07,1.33) {KEEP};
\node[s6, text=sfink, anchor=east] at (5.75,1.08) {angina\_x\_oldpeak};
\draw[fill=sfmid, draw=sfpur, line width=0.4pt, rounded corners=0.0pt] (5.85,1.00) rectangle (6.34,1.17);
\node[s6, text=sfink, anchor=west] at (6.98,1.08) {0.045};
\draw[fill=sfmid, draw=sfpur, line width=0.4pt, rounded corners=1.1pt] (7.60,0.99) rectangle (8.55,1.17);
\node[b55, text=sfink] at (8.07,1.08) {REFINE};
\node[s6, text=sfink, anchor=east] at (5.75,0.83) {flat\_slope\_high\_oldpeak};
\draw[fill=sflt, draw=sfpur, line width=0.4pt, rounded corners=0.0pt] (5.85,0.75) rectangle (5.87,0.92);
\node[s6, text=sfink, anchor=west] at (6.98,0.83) {0.001};
\draw[fill=sflt, draw=sfpur, line width=0.4pt, rounded corners=1.1pt] (7.60,0.74) rectangle (8.55,0.92);
\node[b55, text=sfink] at (8.07,0.83) {REPLACE};
\node[s6, text=sfink, anchor=west] at (3.10,0.50) {Hint: combine Cholesterol and MaxHR};
\node[s6, text=sfink, anchor=west] at (3.10,0.26) {Flag: male\_risk has $r=1.00$ with Sex};
\draw[fill=sfgrey, draw=sfgline, line width=0.4pt, rounded corners=0.0pt] (0.00,1.71) rectangle (0.14,1.85);
\node[s6, text=sfink, anchor=west] at (0.22,1.78) {LLM-Evolutionary Strategy};
\draw[fill=sfpurfill, draw=sfpur, line width=0.4pt, rounded corners=0.0pt] (0.00,1.46) rectangle (0.14,1.60);
\node[s6, text=sfink, anchor=west] at (0.22,1.53) {SHAP feedback (new)};
\draw[fill=sfpur, draw=sfpur, line width=0.4pt, rounded corners=0.0pt] (0.00,1.13) rectangle (0.14,1.27);
\node[s6, text=sfink, anchor=west] at (0.20,1.20) {keep};
\draw[fill=sfmid, draw=sfpur, line width=0.4pt, rounded corners=0.0pt] (0.75,1.13) rectangle (0.89,1.27);
\node[s6, text=sfink, anchor=west] at (0.95,1.20) {refine};
\draw[fill=sflt, draw=sfpur, line width=0.4pt, rounded corners=0.0pt] (1.60,1.13) rectangle (1.74,1.27);
\node[s6, text=sfink, anchor=west] at (1.80,1.20) {replace};
\node[s6, text=sfink, anchor=west] at (0.00,0.75) {B = 200 programs per fold};
\node[s6, text=sfink, anchor=west] at (0.00,0.50) {$\mu$ = $\lambda$ = 5, one child each};
\end{tikzpicture}}
\caption{\shapevo{}. Grey blocks are parts of the $(\mu+\lambda)$ evolution strategy, purple blocks are the newly proposed SHAP-based report. TreeSHAP of the fitted XGBoost model produces a keep/refine/replace report with interaction hints and redundancy flags, which enters the next mutation prompt. The inset shows an excerpt of such a report.}
\label{fig:overview}
\end{figure}

\shapevo{} (Fig.~\ref{fig:overview}) implements this idea. \shapevo{} uses a $(\mu{+}\lambda)$ evolution strategy~\cite{beyer2002es} building on the open source LLaMEA~\cite{vanstein2024llamea}, an LLM-driven evolutionary framework first proposed for automatic metaheuristic design. LLaMEA's mutation prompt carries the parent's error trace and metrics. In feature engineering the counterpart of that feedback is the explanation of the fitted model, which the evaluation already produces. After each evaluation, TreeSHAP values~\cite{lundberg2017shap,lundberg2020treeshap} of the XGBoost model~\cite{chen2016xgboost} trained on the transformed data are summarised in a SHAP report, which goes into the prompt of the next iteration.

Published comparisons of LLM-based feature engineering often differ in backbone, budget, split protocol and leakage regime, for example CAAFE~\cite{hollmann2024caafe} reports GPT-4 with TabPFN over ten proposals while \llmfe~\cite{abhyankar2025llmfe} reports GPT-3.5 with XGBoost over 200 programs, so a reported gain cannot be attributed to the search algorithm. We define a benchmarking protocol in which both methods share the backbone, the budget, the outer cross-validation splits and \llmfe's own evaluation script, and ran it with three backbones on 22 datasets. We find the following.

\begin{itemize}
\item \textbf{SHAP guidance makes the search cheaper with a capable backbone.} With DeepSeek and Haiku, \shapevo{} reaches \llmfe's final validation level with 38 and 62 programs on the median dataset, compared to about 180 for \llmfe, in a third to a half of the wall-clock time and with about a third of the LLM tokens. Its normalised area under the best-so-far curve is larger on 20 and 17 of 22 datasets (Wilcoxon $p<0.001$ and $p=0.003$). With GPT-3.5 it trails (Sect.~\ref{sec:anytime}).
\item \textbf{The SHAP report works when the backbone follows it.} The same loop without the report needs about twice as many programs with DeepSeek and Haiku. The saved prompts show why. DeepSeek and Haiku create again far more of the parent features the report marks \textsc{keep} than of those it marks \textsc{replace}, GPT-3.5 creates both again at the same rate, and without the report all three backbones treat every label alike, so the difference is caused by the report (Sect.~\ref{sec:noshap}).
\item \textbf{A matched comparison with parity on held-out data.} With the search algorithm as the only difference, \shapevo{} is on par with \llmfe{} on held-out accuracy with all three backbones (Sect.~\ref{sec:h2h}).
\end{itemize}

With a capable backbone, SHAP guidance reaches a similar or better solution with significantly fewer evaluations, tokens and time. With a backbone that ignores the report it is harmless, and on no backbone is \llmfe{} the better method on held-out data.

%======================================================================
\section{Related Work}
\label{sec:related}

\paragraph{Automated feature engineering.} Classical approaches enumerate or search a fixed space of transformations, such as deep feature synthesis~\cite{kanter2015deep}, iterative sub-sampling in \texttt{autofeat}~\cite{horn2020autofeat}, reinforcement learning over transformation graphs~\cite{khurana2018feature} and feature boosting with pruning in OpenFE~\cite{zhang2023openfe}. They have no access to domain knowledge and cannot express arbitrary code.

\paragraph{LLM-based feature engineering.} CAAFE~\cite{hollmann2024caafe} prompts an LLM with the dataset description to propose features one at a time and keeps those that improve validation accuracy. FeatLLM~\cite{han2024featllm} extracts rules as binary features for few-shot learning, and OCTree~\cite{nam2024octree} uses decision-tree reasoning to guide iterative refinement. \llmfe~\cite{abhyankar2025llmfe} casts the problem as evolutionary program search. Its experience buffer of programs is partitioned into islands, prompts are built from the best programs of one island, three programs are sampled per prompt, and the reported numbers are an ensemble of the best program of each island. None of these methods feeds feature-level attributions back to the LLM.

\paragraph{LLMs as evolutionary optimizers.} FunSearch~\cite{romeraparedes2024funsearch} and LLaMEA~\cite{vanstein2024llamea} showed that an LLM inside an evolutionary loop can discover programs that outperform hand-designed ones. LLaMEA uses a compact $(\mu{+}\lambda)$ strategy with textual feedback and a self-debugging step, and a recent extension, LLaMEA-SAGE, turns explanations of a surrogate model over syntax-tree features of the generated code into natural-language mutation instructions~\cite{yang2025evostar_cfg}. We move the idea from \emph{code} features to \emph{data} features, since our guidance comes from attributions of the trained model on the engineered columns.

\paragraph{Explanations in the loop.} SHAP~\cite{lundberg2017shap,lundberg2020treeshap} is widely used for post-hoc interpretation and feature selection but rarely as an online search signal. To our knowledge this is the first work to turn per-feature and interaction attributions into natural-language mutation guidance for an LLM feature-engineering search and to evaluate the effect under a fully matched protocol.

%======================================================================
\section{Method}
\label{sec:method}

\subsubsection*{Problem Setting}
Given a tabular dataset $\mathcal{D}=\{(\mathbf{x}_i,y_i)\}_{i=1}^{N}$ with $p$ named columns and a short natural-language description $\mathcal{M}$, we search for a program $T$ implementing \texttt{modify\_features(df)} that maps a data frame to an augmented data frame, such that a fixed learner $f$ (XGBoost with default hyper-parameters) trained on $T(\mathcal{D}_{\mathrm{train}})$ performs best on unseen rows. During the search, fitness is the mean 4-fold inner cross-validation score of $f$ on the training portion of an outer fold. The held-out outer fold is used only for the final evaluation.

\subsubsection*{$(\mu{+}\lambda)$ Evolution Strategy}
\shapevo{} maintains $\mu=5$ programs and produces $\lambda=5$ offspring per generation until a budget of $B=200$ evaluations is spent (Algorithm~\ref{alg:shapevo}). The population is initialised with the identity program and $\mu-1$ programs obtained by prompting the LLM with the task description. Each generation, every survivor is the parent of exactly one offspring, so there is no parent selection step. All selection pressure comes from the survivor step, which keeps the best $\mu$ of parents and offspring after removing identical code. The LLM is sampled at temperature $0.8$ with the completion limit shared with \llmfe{}, 1024 or 2048 tokens depending on the backbone (Sect.~\ref{sec:setup}). Offspring that fail to execute or produce non-finite values are discarded. The failure and its error message are logged, and the most recent failures are shown in later prompts.

\begin{algorithm}[t]
\caption{\shapevo}
\label{alg:shapevo}
\begin{algorithmic}[1]
\REQUIRE description $\mathcal{M}$, training rows $\mathcal{D}_{\mathrm{tr}}$, LLM $\pi$, budget $B$, $\mu$, $\lambda$
\STATE $P \gets \{\text{identity}\}\cup\{\pi(\mathcal{M})\}_{1..\mu-1}$; evaluate each $p\in P$ and build its SHAP report $R_p$
\WHILE{evaluations $<B$}
  \FOR{each parent $p\in P$ (one offspring per survivor)}
    \STATE $c_p \gets \pi(\text{Prompt}(\mathcal{M}, p, R_p, \text{scores of } P))$
    \STATE evaluate $c_p$ by inner CV; TreeSHAP of the fitted model $\to R_{c_p}$
  \ENDFOR
  \STATE $P \gets$ best $\mu$ of $P\cup\{c_p\}$ with distinct code
\ENDWHILE
\RETURN all evaluated programs with fitness (top-3 are ensembled for reporting)
\end{algorithmic}
\end{algorithm}

\subsubsection*{SHAP Feedback Report}
\label{sec:shapreport}
After a candidate is evaluated, the XGBoost model of the last inner fold is explained with TreeSHAP. Mean absolute SHAP values are normalised to sum to one over all columns and every \emph{generated} feature is labelled \textsc{keep} if its share is at least $0.05$, \textsc{refine} if it is at least $0.01$ and below $0.05$, and \textsc{replace} if it is below $0.01$. Because the shares sum to one, a program with many columns pushes more of its features below the thresholds, so large programs receive more  \textsc{refine} and \textsc{replace} labels than small ones. The SHAP report has two additional parts: TreeSHAP also measures how strongly two features act together. The five strongest pairs, prioritizing those involving original columns, are included in the report as \emph{interaction hints}, prompting the LLM to construct features that combine the two variables in each pair. A generated feature whose Pearson correlation with an original column is above $0.9$ is little more than a renamed copy of that column, so the report flags it as \emph{redundant}. In the prompt the report is a list of short lines, one per generated feature, such as ``\texttt{bp\_hr\_ratio: SHAP importance = 0.0040 [LOW - REPLACE]}'', followed by one sentence of instructions for each label, each interaction hint and each redundancy flag. Figure~\ref{fig:example} in Sect.~\ref{sec:noshap} shows a complete report from one run. The rest of the prompt contains the standard elements: the task description, the parent’s code and fitness, the fitness of five population members, and a row-wise constraint prohibiting group-by statistics, ranks, frequencies, and other cross-row operations (Sect.~\ref{sec:h2h}).

%======================================================================
\section{Experimental Setup}
\label{sec:setup}

\subsection{Datasets and Backbones}
\label{sec:datasets}
We use 21 of the 22 datasets of the \llmfe{} benchmark for which data and task descriptions were released, 16 classification and 5 regression tasks with 625 to 48{,}842 rows (Table~\ref{tab:datasets} lists sizes). We leave out \texttt{forest-fires}, on which the published \llmfe{} score could not be reproduced because 48\% of its targets are zero and a few extreme values dominate the RMSE under any split (Table~\ref{tab:repro}). We add one \textsc{synthetic} binary-classification dataset generated for this study, ensuring it was unseen by LLMs during pretraining.  Since its rows are grouped, all its splits keep a group's rows together with scikit-learn's \texttt{StratifiedGroupKFold}: the 5 outer folds, which are also the folds of the final evaluation, and the 4 inner folds of both searches. We test both methods with three backbones, \texttt{gpt-3.5-turbo-0125}, the backbone of the original \llmfe{} paper, \texttt{deepseek-v4-flash}, a low-cost open-weight model, and \texttt{claude-haiku-5.5}, a recent efficient model from Anthropic. All models are accessed through OpenRouter with reasoning disabled. Both methods use the same token limits: 1024 for GPT-3.5 and 2048 for DeepSeek and Haiku, which generate longer programs. Each backbone requires 22{,}000 program evaluations per method (22 datasets × 5 folds × 200 programs). 

\subsection{Protocol}
\label{sec:protocol}
Table~\ref{tab:hyper} summarises both methods. Every experiment uses 5-fold outer cross-validation with seed 42. Both methods receive exactly the same training rows in every fold and the same budget of 200 program evaluations per fold, the setting of the \llmfe{} paper. Inside a fold, both methods score candidates by 4-fold inner cross-validation of XGBoost with fixed seed (42), using accuracy or negative RMSE for regression. Every LLM call is charged one budget unit in both methods, whether or not it yields a program that runs.  \llmfe{} runs from the authors' public code through OpenRouter, with its completion limit raised from the authors' default of 512 tokens to the shared limit of Sect.~\ref{sec:datasets}, and both extractors repair replies cut off before the \texttt{return} statement in the same way, so that neither method loses programs to truncation. The remaining extractor details, which concern program handling and not the search, are in Appendix~\ref{sec:harness}. \llmfe{} samples at the API's default temperature of $1.0$, as in the authors' code, \shapevo{} at $0.8$. To separate the contribution of the explanation signal from that of the evolution strategy we also run a \emph{no-SHAP control}, which is the same \shapevo{} code, splits, budget and completion limit, but with the SHAP report, interaction hints and redundancy flags removed from the prompt and SHAP not computed, so the LLM sees only code and scores as in \llmfe. 

Final scores are computed with \llmfe's official evaluation script. The three highest-scoring programs of a fold are applied to the held-out rows, one XGBoost model with seed 42 is trained per program, and the predictions are ensembled by majority vote over the predicted labels for classification and by their mean for regression. We apply the same script to \shapevo's three best programs. Regression is reported as N-RMSE $=$ RMSE$/\mathrm{mean}(y)$. 

\begin{table}[t]
\centering
\caption{Search settings of the two methods. Shared settings are given in the text.}
\label{tab:hyper}
\footnotesize
 \begin{tabular}{@{}>{\raggedright\arraybackslash}p{0.15\textwidth}>{\raggedright\arraybackslash}p{0.38\textwidth}>{\raggedright\arraybackslash}p{0.39\textwidth}@{}}
\toprule
 & LLM-FE & SHAPEvo-FE \\
\midrule
Search & island buffer (3 islands), 2 examples/prompt, 3 samples/prompt & $(\mu{+}\lambda)$-ES, $\mu=\lambda=5$, 1 sample/prompt \\
Selection & cluster-softmax sampling of prompt examples & every survivor mutated once per generation; elitist $(\mu{+}\lambda)$ survivors \\
Feedback in prompt & code + score of examples & code + score + SHAP report (keep/refine/replace, interactions, redundancy) \\
LLM sampling & API default temperature; $\le 1024$ completion tokens (GPT-3.5), $\le 2048$ (DeepSeek, Haiku) & $T=0.8$; same completion limits \\
Allowed operators & row-wise and cross-row (group-by, rank, frequency) & row-wise only \\
\bottomrule
\end{tabular}

\end{table}
All \llmfe{} numbers in this paper are from our re-run of the authors' code under the shared protocol. The re-run matches the published table to a median absolute difference of $0.004$ (Table~\ref{tab:repro}), and the evaluator details that affect fairness are documented in Appendix~\ref{sec:harness}. Since a single seed is used per backbone, a difference counts as a win only if $|\Delta|>0.01$, which is well inside the fold-to-fold standard deviation of either method.

%======================================================================
\section{Results}
\label{sec:results}

\subsection{Held-Out Performance}
\label{sec:h2h}
The per-dataset held-out scores are in Table~\ref{tab:h2h} and their wins, ties and losses in Table~\ref{tab:summary}. \shapevo{} achieves comparable held-out performance to \llmfe{} with GPT-3.5 and a modest advantage with DeepSeek and Haiku. Most datasets are ties under the chosen threshold. Improvements on \texttt{vehicle} and \texttt{housing} recur with both DeepSeek and Haiku, where the generated programs use ratios of silhouette measurements and per-room or per-household quantities. The advantage is not universal, however. \llmfe{} performs better on \texttt{cmc} with both backbones, and the methods tie on \textsc{synthetic} with all three.

Some of the largest gains arise from failures in \llmfe's generated programs. On \texttt{car} and \texttt{bike}, group-by statistics change when computed separately on training and test rows, causing poor predictions on some held-out folds. With Haiku, an index-alignment error on one \texttt{tic-tac-toe} fold corrupts the generated features after the program removes the original columns. On the Haiku \texttt{tic-tac-toe} and \texttt{car} folds the collapse depends on how the official scorer breaks ties between programs with the same inner-CV score (Appendix~\ref{sec:h2h_supp}). These failures account for part of the observed advantage and should be distinguished from improvements in feature quality. \shapevo's row-wise restriction avoids the cross-row failure, but also narrows the available transformations. A separate audit on six categorical datasets found little average benefit from computing statistics jointly on training and test rows, so leakage does not explain \llmfe's scores.

With GPT-3.5, \llmfe's prompt for \texttt{communities} exceeds the context window and only one fold completes, while \shapevo{} uses fewer example rows for wide datasets and completes all folds, so we exclude \texttt{communities} from the GPT-3.5 comparisons. Both methods complete all folds with DeepSeek and Haiku.

At the full budget both searches approach a plateau on the inner objective for most datasets, so the main differences concern how quickly they reach comparable validation scores and whether the backbone uses the SHAP feedback, as examined in Sects.~\ref{sec:anytime} and~\ref{sec:noshap}.

\subsection{Search Speed Depends on the Backbone}
\label{sec:anytime}

\begin{figure}[tbp]
\centering
\includegraphics[width=\textwidth]{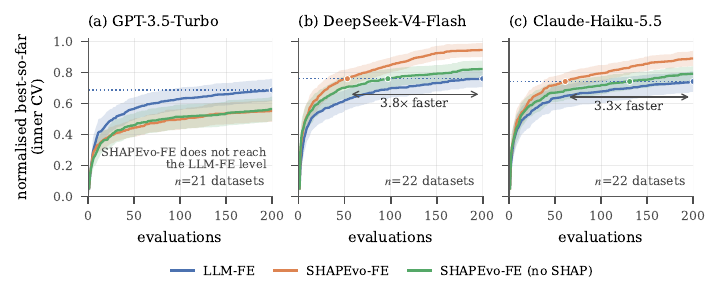}
\caption{Search speed with GPT-3.5-Turbo (a), DeepSeek-V4-Flash (b) and Claude-Haiku-5.5 (c), all datasets. Mean rescaled best-so-far inner-CV score over datasets with a 95\% confidence band, for \llmfe, \shapevo{} and \shapevo{} with the SHAP report removed from the prompt (Sect.~\ref{sec:noshap}). The dotted line is \llmfe's level at the end of its budget, the dots mark where each mean curve reaches it, and the arrow gives the ratio of the two positions. The speed-up at equal level along the whole budget is in Supplementary Fig.~S1.}
\label{fig:convergence}
\end{figure}

Held-out parity leaves open how each method spends its budget. Both searches log the best inner-CV score after every evaluation on the same inner folds, so the curves are directly comparable. Figure~\ref{fig:convergence} shows them together with the no-SHAP control of Sect.~\ref{sec:noshap}. For every dataset the best-so-far inner-CV score is averaged over the five folds and rescaled to $[0,1]$ between the worst starting score and the best final score of its nine curves (three methods, three backbones), so the three panels share one scale. The rescaling gives every dataset the same weight, so the gap between the curves ranks the methods and does not measure an accuracy difference. The x-axis counts evaluated programs, that is, LLM calls.

Table~\ref{tab:summary} reads the same curves without any rescaling. For every dataset it takes the point at which the fold-mean curve of each method first reaches \llmfe's final fold-mean level on that dataset, in programs, in wall-clock minutes and in LLM tokens, and reports the median over datasets. DeepSeek provides the strongest evidence for \shapevo. Its mean curve stays above \llmfe's for 99\% of the budget range, and on the median dataset \shapevo{} reaches \llmfe's final level with 38 programs where \llmfe{} needs 180, in 24 instead of 66 minutes and with 332k instead of 950k tokens. It gets there first on 20 of 22 datasets. Haiku repeats the picture with 62 instead of 186 programs, 9 instead of 20 minutes and 257k instead of 785k tokens, first on 19 of 22 datasets. The token advantage is smaller than the program advantage because a \shapevo{} prompt carries the SHAP report and is longer, 9.1k against 6.3k tokens per call with DeepSeek, and the time advantage is smaller because every \shapevo{} evaluation also computes TreeSHAP. The advantage is not due to a bigger portion of valid programs, since the median number of scored programs per fold differs by at most 6\% between the two methods with DeepSeek and Haiku (Table~\ref{tab:audit}). Table~\ref{tab:auc} summarises every curve by its normalised area, the mean of the rescaled best-so-far curve over the 200 evaluations, so that 1 would mean sitting at the best final level from the first evaluation on. \shapevo{} has the larger area on 20 of 22 datasets with DeepSeek and 17 of 22 with Haiku (two-sided Wilcoxon signed-rank test over the paired datasets, $p<0.001$ and $p=0.003$) and the smaller one on 19 of 21 with GPT-3.5 ($p<0.001$).

\begin{table}[t]
\centering
\caption{Normalised area under the best-so-far inner-CV curve over 200 evaluations, mean over the five folds, rescaled per dataset and backbone between the worst start and the best final value of the three arms (\llmfe, \shapevo, no-SHAP control of Sect.~\ref{sec:noshap}). $\Delta>0$ favours \shapevo. Bold marks the larger area. $p$ is the two-sided Wilcoxon signed-rank test over the paired per-dataset values. With GPT-3.5 \texttt{communities} is left out because \llmfe{} completed one fold only. $^{R}$ marks regression.}
\label{tab:auc}
\scriptsize
\resizebox{\textwidth}{!}{\begin{tabular}{lrrrrrrrrr}
\toprule
Dataset & \multicolumn{3}{c}{GPT-3.5-Turbo} & \multicolumn{3}{c}{DeepSeek-V4-Flash} & \multicolumn{3}{c}{Claude-Haiku-5.5} \\
 & \llmfe & \shapevo & $\Delta$ & \llmfe & \shapevo & $\Delta$ & \llmfe & \shapevo & $\Delta$ \\
\midrule
adult & \textbf{0.837} & 0.762 & -0.075 & \textbf{0.529} & 0.493 & -0.037 & \textbf{0.913} & 0.819 & -0.095 \\
balance-scale & \textbf{0.958} & 0.897 & -0.061 & 0.992 & \textbf{0.995} & +0.003 & 0.995 & \textbf{0.995} & +0.000 \\
bank & \textbf{0.825} & 0.759 & -0.066 & 0.733 & \textbf{0.902} & +0.169 & 0.422 & \textbf{0.797} & +0.375 \\
blood & \textbf{0.799} & 0.527 & -0.272 & 0.666 & \textbf{0.782} & +0.116 & 0.620 & \textbf{0.836} & +0.216 \\
breast-w & \textbf{0.868} & 0.754 & -0.114 & 0.725 & \textbf{0.852} & +0.126 & 0.776 & \textbf{0.829} & +0.053 \\
car & \textbf{0.860} & 0.656 & -0.204 & 0.616 & \textbf{0.957} & +0.341 & 0.812 & \textbf{0.902} & +0.091 \\
cmc & \textbf{0.805} & 0.564 & -0.241 & 0.537 & \textbf{0.737} & +0.200 & \textbf{0.777} & 0.741 & -0.036 \\
communities & -- & -- & -- & 0.680 & \textbf{0.837} & +0.157 & \textbf{0.806} & 0.803 & -0.003 \\
credit-g & 0.774 & \textbf{0.845} & +0.071 & 0.598 & \textbf{0.791} & +0.193 & 0.645 & \textbf{0.769} & +0.124 \\
eucalyptus & \textbf{0.834} & 0.623 & -0.211 & 0.568 & \textbf{0.836} & +0.268 & 0.493 & \textbf{0.858} & +0.365 \\
heart & \textbf{0.785} & 0.624 & -0.161 & 0.556 & \textbf{0.858} & +0.302 & \textbf{0.754} & 0.748 & -0.006 \\
junglechess & \textbf{0.957} & 0.609 & -0.348 & 0.965 & \textbf{0.990} & +0.025 & 0.951 & \textbf{0.957} & +0.006 \\
myocardial & \textbf{0.831} & 0.651 & -0.180 & 0.677 & \textbf{0.814} & +0.137 & 0.490 & \textbf{0.873} & +0.383 \\
pc1 & \textbf{0.919} & 0.313 & -0.606 & 0.623 & \textbf{0.812} & +0.188 & 0.747 & \textbf{0.856} & +0.110 \\
synthetic & \textbf{0.795} & 0.752 & -0.042 & 0.649 & \textbf{0.842} & +0.194 & 0.658 & \textbf{0.829} & +0.172 \\
tic-tac-toe & \textbf{0.851} & 0.737 & -0.114 & 0.427 & \textbf{0.827} & +0.400 & 0.446 & \textbf{0.837} & +0.391 \\
vehicle & \textbf{0.889} & 0.597 & -0.292 & 0.607 & \textbf{0.842} & +0.235 & 0.509 & \textbf{0.797} & +0.288 \\
bike$^{R}$ & 0.537 & \textbf{0.876} & +0.339 & \textbf{0.720} & 0.702 & -0.018 & 0.740 & \textbf{0.785} & +0.045 \\
crab$^{R}$ & \textbf{0.883} & 0.655 & -0.228 & 0.830 & \textbf{0.880} & +0.050 & \textbf{0.882} & 0.673 & -0.209 \\
housing$^{R}$ & \textbf{0.827} & 0.612 & -0.215 & 0.871 & \textbf{0.933} & +0.062 & 0.873 & \textbf{0.924} & +0.051 \\
insurance$^{R}$ & \textbf{0.797} & 0.639 & -0.158 & 0.702 & \textbf{0.895} & +0.192 & 0.632 & \textbf{0.832} & +0.200 \\
wine$^{R}$ & \textbf{0.915} & 0.805 & -0.111 & 0.637 & \textbf{0.858} & +0.220 & 0.719 & \textbf{0.874} & +0.155 \\
\midrule
mean & 0.836 & 0.679 & -0.157 & 0.678 & 0.838 & +0.160 & 0.712 & 0.833 & +0.122 \\
\shapevo{} higher on, Wilcoxon $p$ & \multicolumn{3}{c}{2 of 21, $p<0.001$} & \multicolumn{3}{c}{20 of 22, $p<0.001$} & \multicolumn{3}{c}{17 of 22, $p=0.003$} \\
no-SHAP control, mean & \multicolumn{3}{c}{0.687} & \multicolumn{3}{c}{0.747} & \multicolumn{3}{c}{0.759} \\
Wilcoxon $p$, \shapevo{} vs control, control vs \llmfe & \multicolumn{3}{c}{$p=0.812$, $p<0.001$} & \multicolumn{3}{c}{$p=0.005$, $p=0.007$} & \multicolumn{3}{c}{$p=0.017$, $p=0.033$} \\
\bottomrule
\end{tabular}
}
\end{table}

GPT-3.5 shows the opposite pattern. \llmfe{} leads for 99\% of the budget range and \shapevo{} reaches its final level on one dataset only. Removing the SHAP report does not change this (Sect.~\ref{sec:noshap}), so the difference comes from the search strategy. \llmfe{} shows GPT-3.5 two strong programs and asks it to write a third, while \shapevo{} provides one parent and a SHAP report, and GPT-3.5 appears to benefit more from combining programs than from feature feedback. With every backbone the final gaps on the search objective are small, a mean accuracy gap of 0.006 on the classification datasets in favour of \shapevo{} with DeepSeek and of \llmfe{} with GPT-3.5, consistent with the held-out ties.

\begin{table}[t]
\centering
\caption{Cost of reaching \llmfe's final inner-CV level. For each dataset the target is \llmfe's fold-mean score after 200 programs. The table gives how much each method spends before its own fold-mean curve first reaches that target, as \shapevo{} / \llmfe: programs, wall-clock minutes and thousands of LLM tokens (prompt and reply, one tokenizer for both methods, Appendix~\ref{sec:convergence_supp}). \llmfe{} reaches its own final level on every dataset, \shapevo{} does not. The bracket counts the datasets on which it does, and its entries are medians over those datasets. \llmfe's entries are medians over all datasets. With GPT-3.5 \shapevo{} reaches the target on one dataset only, marked by a dash. The last column counts the datasets on which \shapevo{} needs fewer programs than \llmfe. W/T/L are the held-out wins, ties and losses of Sect.~\ref{sec:h2h} (Table~\ref{tab:h2h}).}

\label{tab:summary}
\scriptsize
\setlength{\tabcolsep}{4pt}
\begin{tabular}{lccccc}
\toprule
& held-out & \multicolumn{3}{c}{to \llmfe's final level, \shapevo{} / \llmfe} & faster on \\
\cmidrule(lr){3-5}
Backbone & W/T/L & programs & minutes & k tokens & datasets \\
\midrule
GPT-3.5-Turbo & 2/18/2 & -- / 184 (1 of 21) & -- / 21 & -- / 475 & 1 of 21 \\
DeepSeek-V4-Flash & 6/13/3 & 38 / 180 (20 of 22) & 24 / 66 & 332 / 950 & 20 of 22 \\
Claude-Haiku-5.5 & 6/15/1 & 62 / 186 (20 of 22) & 9 / 20 & 257 / 785 & 19 of 22 \\
\bottomrule
\end{tabular}
\end{table}

\subsection{Explanation or Strategy? The No-SHAP Control}
\label{sec:noshap}
To analyse the contribution to the convergence speed advantage of Sect.~\ref{sec:anytime} to either the $(\mu{+}\lambda)$ strategy or the SHAP report we ran a no-SHAP control (Sect.~\ref{sec:protocol}). It runs the same code, splits and budget with the SHAP report, interaction hints and redundancy flags removed from the prompt, so the LLM sees only code and scores as in \llmfe. Figure~\ref{fig:convergence} compares the three arms on the three backbones. With DeepSeek both ingredients contribute, the SHAP report more. The control alone is above \llmfe{} over the whole budget and reaches \llmfe's final level on the mean curve after 97 programs instead of 200, a gain that comes from the elitist $(\mu{+}\lambda)$ loop with the parent's code, score and errors in the prompt. The SHAP report brings this to 53 programs. The full method is above the control on 97\% of the budget and reaches the common score first on 18 of 22 datasets, with a median speed-up of $2.4\times$. Held-out accuracy moves as well. Full and control differ by more than $0.01$ on six datasets, four in favour of the SHAP report and two against, and the control alone is on par with \llmfe. Haiku repeats this picture. The control needs 131 programs and the full method 61, the full method is above the control on 98\% of the budget and first on 16 of 22 datasets, and on held-out data the SHAP report wins five datasets and loses two. The last rows of Table~\ref{tab:auc} test this over datasets. With DeepSeek and Haiku the area under the control's curve lies between the two methods, significantly above \llmfe{} ($p=0.007$ and $p=0.033$) and significantly below the full method ($p=0.005$ and $p=0.017$). With 
GPT-3.5 the SHAP report makes no difference. Full method and control reach the common score first on 10 and 11 datasets, their areas do not differ ($p=0.81$), neither reaches \llmfe's final level, and held-out accuracy does not favour the report either, with 1 win and 3 losses. Per-dataset held-out scores of all three arms are in Tables~\ref{tab:noshap_gpt35} to~\ref{tab:noshap_haiku}.

\begin{figure}[t]
\centering
\includegraphics[width=\textwidth]{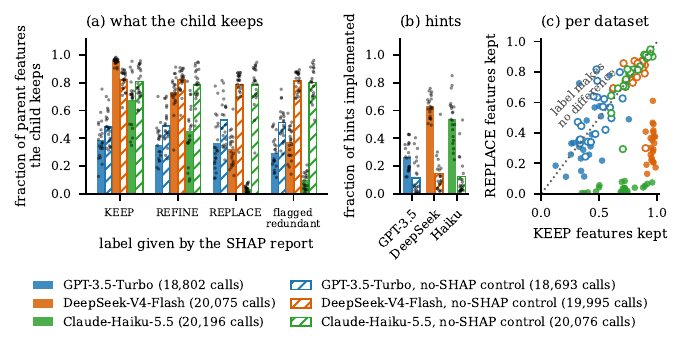}
\caption{Reading the SHAP report: what the child program keeps of the parent's features. Parsed from every mutation prompt that carried a report and the program returned for it (18,802 GPT-3.5 calls, 20,075 DeepSeek calls, 20,196 Haiku calls). (a)~Fraction of the parent's generated features that the child program creates again, grouped by the label the report gave them (\textsc{keep}, \textsc{refine}, \textsc{replace}) and for features the report flagged as redundant. Bars are means over calls, dots are datasets. (b)~Fraction of the report's interaction hints that the child implements by putting both named columns into one new feature. (c)~One point per dataset, with the fraction of \textsc{keep} features kept on the x-axis and the fraction of \textsc{replace} features kept on the y-axis. The dotted diagonal is where the two fractions are equal. Hatched bars and hollow points show the no-SHAP control runs, for which the labels were recomputed after the fact with the search evaluator and never shown to the LLM.}
\label{fig:compliance}
\end{figure}

The saved prompts show why (Fig.~\ref{fig:compliance}). For every mutation prompt that carried a SHAP report we read the label of each of the parent's generated features and checked whether the child program creates that feature again. Over the 20,100 DeepSeek prompts the child creates again 96\% of the features labelled \textsc{keep}, 74\% of those labelled \textsc{refine} and 32\% of those labelled \textsc{replace}, and it realises 63\% of the interaction hints. Over the 20,200 Haiku prompts the three rates are 68\%, 45\% and 3.5\%, with 54\% of the hints realised, so Haiku drops almost every feature the report marks \textsc{replace}. Over the 18,800 GPT-3.5 prompts the three rates are 39\%, 35\% and 36\%, and 26\% of the hints are realised, so the label makes no difference to what GPT-3.5 keeps. Fig.~\ref{fig:compliance}(c) shows the same comparison dataset by dataset, with the fraction of \textsc{keep} features kept on the x-axis and the fraction of \textsc{replace} features kept on the y-axis. Every GPT-3.5 dataset lies on the diagonal, where \textsc{keep} and \textsc{replace} features are kept at the same rate, and every DeepSeek and Haiku dataset lies far below it.

Figure~\ref{fig:example} shows what following the report looks like in one call. The parent has ten generated features, the report labels four \textsc{keep}, three \textsc{refine} and three \textsc{replace}, and it asks for three interactions. The child keeps the four \textsc{keep} features line for line, omits the three \textsc{replace} features, keeps the three \textsc{refine} features and adds one product feature for each of the two hinted pairs it did not have yet. Its inner-CV accuracy rises from $0.866$ to $0.873$. The two new features are the only part of the program the LLM had to invent, and the report told it which columns to combine.

A strong backbone might keep high-SHAP features simply because they look useful, label or no label. The control runs rule this out. For every mutation prompt of the control runs we recomputed the parent's SHAP report post hoc with the search evaluator, which reproduces the parent's printed score in every case, and scored the child in the same way. The LLM never saw these labels. Without the report DeepSeek creates again 82\% of the features that would have been labelled \textsc{keep} and 79\% of those that would have been labelled \textsc{replace}, Haiku 81\% and 79\%, and GPT-3.5 48\% and 53\% (hatched bars and hollow points in Fig.~\ref{fig:compliance}). All three backbones treat the labels alike when they cannot see them, so the gap between \textsc{keep} and \textsc{replace} in the full runs is caused by the report. The report changes what DeepSeek and Haiku keep, and these are the backbones on which it halves the number of programs needed. GPT-3.5 ignores it, and with GPT-3.5 it has no effect.

\begin{figure}[t]
{\scriptsize
\begin{verbatim}
=== Feature Impact Report ===               (parent score 0.866)
Generated features (by your function):
  hr_deficit_x_oldpeak:    SHAP importance = 0.0695 [HIGH - KEEP]
  ecg_x_slope:             SHAP importance = 0.0644 [HIGH - KEEP]
  chol_slope_score:        SHAP importance = 0.0526 [HIGH - KEEP]
  cp_chol_product:         SHAP importance = 0.0893 [HIGH - KEEP]
  oldpeak_x_slope:         SHAP importance = 0.0070 [LOW - REPLACE]
  angina_x_oldpeak:        SHAP importance = 0.0448 [MEDIUM - REFINE]
  slope_x_hr_deficit:      SHAP importance = 0.0146 [MEDIUM - REFINE]
  ecg_slope_oldpeak:       SHAP importance = 0.0242 [MEDIUM - REFINE]
  flat_slope_high_oldpeak: SHAP importance = 0.0011 [LOW - REPLACE]
  high_chol_angina:        SHAP importance = 0.0016 [LOW - REPLACE]
Try to capture the interaction between Cholesterol and MaxHR
Try to capture the interaction between ST_Slope and hr_deficit_x_oldpeak
Try to capture the interaction between Sex and ST_Slope

=== Child program ===                       (child score 0.873)
  hr_deficit_x_oldpeak = (220 - Age - MaxHR) * Oldpeak           kept
  ecg_x_slope          = RestingECG * (2 - ST_Slope)             kept
  chol_slope_score     = Cholesterol * (2 - ST_Slope) / 100      kept
  cp_chol_product      = ChestPainType * Cholesterol / 100       kept
  chol_x_maxhr_norm    = Cholesterol * MaxHR / 10000             new, hint 1
  slope_x_hr_deficit   = (2 - ST_Slope) * hr_deficit_x_oldpeak   refined, hint 2
  sex_x_slope          = Sex * (2 - ST_Slope)                    new, hint 3
  ecg_slope_oldpeak    = RestingECG * (2 - ST_Slope) * Oldpeak   refined
  angina_x_oldpeak     = ExerciseAngina * Oldpeak                refined
\end{verbatim}}
\caption{One SHAP report and the program it produced, from prompt 150 on \texttt{heart}, fold 2, with Claude-Haiku-5.5. The top block is the report as it appears in the prompt, with the three interaction hints of the guidance section (the five original columns still contributing and the instruction sentences are omitted). The bottom block lists the assignments of the child program in the order written, with the original column names in place of the \texttt{df\_input[...]} syntax. The child creates the four \textsc{keep} features unchanged, omits the three \textsc{replace} features, keeps the three \textsc{refine} features, one of which is already the pair of the second hint, and adds one new feature for each of the other two hints. Scores are inner-CV accuracy.}
\label{fig:example}
\end{figure}

\subsection{How Many Features the Programs Add}
\label{sec:parsimony}
Table~\ref{tab:features} counts the new feature columns that the best program of each fold adds, as the median over folds, and how many of them the fitted model still uses. The number added depends mostly on the backbone. With GPT-3.5 both methods add about 3 features. With DeepSeek, \llmfe{} adds 18 and \shapevo{} 46, and with Haiku 14 and 10. This difference in program size is already there before the first SHAP report exists. The initial programs DeepSeek writes for \shapevo{}, which are generated before any program has been evaluated, add a median of 40 features and the best program at the end adds 45 (medians over all folds), whereas GPT-3.5 starts at 1 and ends at 3, so the report does not make the programs grow. 

The SHAP labels push in opposite directions on the two backbones. With GPT-3.5 the \shapevo{} program is smaller than the \llmfe{} program on 13 of the 21 datasets, the same size on 6 and larger on 2, so the labels mainly remove features. With Haiku it is smaller on 14 of 21, the same size on 2 and larger on 5. With DeepSeek, which follows the \textsc{refine} and interaction hints, the \shapevo{} program is larger on 20 of 21 datasets, and these larger programs are the ones with the faster inner-CV progress of Sect.~\ref{sec:anytime}. Yet the absolute threshold retains a median of only 1 to 4 added features. Removing the rest changes the held-out score by at most 0.01 on at least 15 of the 21 datasets for every combination of method and backbone (Table~\ref{tab:features}). This threshold favours smaller programs because attribution shares sum to one across all columns.

The rank-based criterion instead retains the fewest added features covering 80\% of their combined attribution. With GPT-3.5, both methods retain a median of 2 added features. With DeepSeek, the medians are 8 for \llmfe{} and 17.5 for \shapevo{}. After rank-based pruning, the held-out score changes by at most 0.01 on at least 18 of the 21 datasets for every combination of method and backbone. The backbones differ in how many idle features a program carries, not in how many the model uses, and neither a small nor a large program is clearly better on held-out data.

\begin{table}[t]
\centering
\caption{Feature counts of each fold's best program, medians over folds and datasets. The per-dataset counts are in Table~\ref{tab:features_full}. Added counts all new features. Kept counts those with a TreeSHAP share $\geq 0.05$ after refitting XGBoost. Rank-based pruning retains the smallest set of top-ranked added features covering at least 80\% of the total attribution to added features. For each pruning rule, the rows below report the mean change in single-program held-out score across datasets, positive when pruning improves performance, and the number of datasets with $|\Delta|\leq 0.01$. Details are in Table~\ref{tab:prune}.}
\label{tab:features}
\scriptsize
\setlength{\tabcolsep}{4pt}
\resizebox{\textwidth}{!}{\begin{tabular}{lrrrrrrrrrrrr}
\toprule
 & \multicolumn{4}{c}{GPT-3.5-Turbo} & \multicolumn{4}{c}{DeepSeek-V4-Flash} & \multicolumn{4}{c}{Claude-Haiku-5.5} \\
 & \multicolumn{2}{c}{LLM-FE} & \multicolumn{2}{c}{SHAPEvo-FE} & \multicolumn{2}{c}{LLM-FE} & \multicolumn{2}{c}{SHAPEvo-FE} & \multicolumn{2}{c}{LLM-FE} & \multicolumn{2}{c}{SHAPEvo-FE} \\
 & added & kept & added & kept & added & kept & added & kept & added & kept & added & kept \\
\midrule
median & 3 & 1.5 & 3 & 1 & 18 & 3 & 46 & 3 & 14 & 3 & 10 & 4 \\
\midrule
$\Delta$ held-out, kept only & \multicolumn{2}{c}{$+0.0008$} & \multicolumn{2}{c}{$+0.0003$} & \multicolumn{2}{c}{$-0.0039$} & \multicolumn{2}{c}{$-0.0009$} & \multicolumn{2}{c}{$-0.0018$} & \multicolumn{2}{c}{$-0.0047$} \\
datasets with $|\Delta|\le0.01$ & \multicolumn{2}{c}{20/21} & \multicolumn{2}{c}{20/21} & \multicolumn{2}{c}{15/21} & \multicolumn{2}{c}{16/21} & \multicolumn{2}{c}{17/21} & \multicolumn{2}{c}{17/21} \\
\midrule
kept, rank-based (median) & \multicolumn{2}{c}{2} & \multicolumn{2}{c}{2} & \multicolumn{2}{c}{8} & \multicolumn{2}{c}{17.5} & \multicolumn{2}{c}{7} & \multicolumn{2}{c}{5} \\
$\Delta$ held-out, rank-based kept only & \multicolumn{2}{c}{$+0.0005$} & \multicolumn{2}{c}{$+0.0000$} & \multicolumn{2}{c}{$+0.0001$} & \multicolumn{2}{c}{$+0.0004$} & \multicolumn{2}{c}{$-0.0006$} & \multicolumn{2}{c}{$-0.0022$} \\
datasets with $|\Delta|\le0.01$ & \multicolumn{2}{c}{21/21} & \multicolumn{2}{c}{21/21} & \multicolumn{2}{c}{20/21} & \multicolumn{2}{c}{19/21} & \multicolumn{2}{c}{19/21} & \multicolumn{2}{c}{18/21} \\
\bottomrule
\end{tabular}
}
\end{table}

%======================================================================
\section{Discussion and Limitations}
\label{sec:discussion}
Our reading of the three backbones is that a model has to act on a recommendation for it to matter. GPT-3.5 keeps a feature at the same rate whether the SHAP report says keep or replace, whereas DeepSeek keeps three times as many of the keep features as of the replace features and Haiku nineteen times as many (68\% against 3.5\%), and the no-SHAP control shows that this selectivity is a response to the report and not a property of the model. The benchmark itself limits what the speed-up can show. Scoring the same runs after 25, 50, 75 or 100 programs per fold changes the median held-out score by less than $0.002$ for either method, so the held-out score is saturated within the first 25 programs while the search objective keeps rising. Faster progress therefore saves programs, time and tokens without changing accuracy here, and its practical value lies in harder tasks and short budgets. 

Limitations. Parent selection is uniform, since every survivor is mutated once per generation, and we did not vary selection pressure. The two methods also sample at different temperatures, the API default of $1.0$ for \llmfe{}, as in the authors' code, and $0.8$ for \shapevo{}, and we did not test whether matching them changes the picture. Each backbone was run with a single seed, and confidence intervals across seeds are the first item of future work. The ensemble reported for both methods is a post-hoc selection of the three best programs. Neither search optimises for ensemble diversity, and \llmfe's islands give it a natural advantage here. The learner is fixed to XGBoost with default hyper-parameters, as in \llmfe. The report labels use absolute share thresholds, which favour small programs, and a rank-based rule that labels the features carrying a fixed share of the attribution would make the report independent of program size. An LLM call that returns no usable program is charged to the budget in both methods. Such calls, including an 18-minute gateway outage, took under 1\% of the budget and at most 14 evaluations of a fold, 30 on one. The synthetic dataset is not publicly available.

%======================================================================
\section{Conclusion}
\label{sec:conclusion}
We introduced \shapevo{}, an LLM-driven  $(\mu{+}\lambda)$ evolution strategy for feature engineering, including a TreeSHAP report that tells the LLM which generated features to keep, refine or replace. Under a matched protocol on 22 datasets with three backbones, our main finding is that with SHAP guidance and a capable backbone the search needs significantly fewer evaluations, tokens and time to reach a similar or better solution. With DeepSeek and Haiku \shapevo{} reaches \llmfe's final validation level with a fifth to a third of the programs (evaluations), a third to a half of the time and about a third of the tokens, its area under the best-so-far curve is significantly larger, and the larger part of that gain comes from the SHAP report. The final held-out accuracy is on par with \llmfe{} or slightly better. With GPT-3.5, which ignores the report's labels, \llmfe{} is slightly faster. SHAP feedback helps when the backbone follows it and makes no difference when the backbone ignores it. Code, the \llmfe{} reproduction, all prompts, the raw logs and the supplementary material are available at \url{https://doi.org/10.5281/zenodo.23750423}.

%======================================================================
\clearpage
\bibliographystyle{splncs04}
\bibliography{refs}

\bigskip
% ---------------- Supplementary material as appendix ----------------
\setcounter{section}{0}\renewcommand{\thesection}{S\arabic{section}}
\setcounter{table}{0}\renewcommand{\thetable}{S\arabic{table}}
\setcounter{figure}{0}\renewcommand{\thefigure}{S\arabic{figure}}
\setcounter{lstlisting}{0}\renewcommand{\thelstlisting}{S\arabic{lstlisting}}
\section*{Appendix: Supplementary Material}
This appendix reproduces the supplementary material submitted with the paper. Its sections, tables, figures and listings are numbered with the prefix S.
\section{Reproduction of the Published \llmfe{} Results}
\label{sec:repro}
\begin{table}[h]
\centering
\caption{\llmfe{} as published in Tables 2 and 3 of its paper (\texttt{gpt-3.5-turbo}, top-3 island ensemble, mean $\pm$ std over five splits) against our re-run of the authors' code with \texttt{gpt-3.5-turbo-0125} through OpenRouter, seed 42 and 1024 completion tokens, scored with the authors' evaluator. This re-run is the one used in the main paper, and the table only shows that the baseline was reproduced faithfully. The differences come from the stochastic LLM sampling of a single run and possibly from a different model snapshot, since re-scoring the authors' released logs with our evaluator reproduces the published numbers exactly.}
\label{tab:repro}
\scriptsize
\resizebox{\textwidth}{!}{\begin{tabular}{llrrrl}
\toprule
Dataset & Metric & published & our re-run & diff. & note \\
\midrule
adult & Acc. & 0.874$\pm$0.003 & 0.872$\pm$0.004 & -0.002 &  \\
arrhythmia & Acc. & 0.659$\pm$0.018 & -- & -- & authors' code crashes on this dataset \\
balance-scale & Acc. & 0.990$\pm$0.013 & 1.000$\pm$0.000 & +0.010 &  \\
bank & Acc. & 0.907$\pm$0.002 & 0.907$\pm$0.003 & -0.000 &  \\
blood & Acc. & 0.751$\pm$0.036 & 0.741$\pm$0.030 & -0.010 &  \\
breast-w & Acc. & 0.970$\pm$0.009 & 0.957$\pm$0.010 & -0.013 &  \\
car & Acc. & 0.999$\pm$0.001 & 1.000$\pm$0.000 & +0.001 &  \\
cdc-diabetes & Acc. & 0.849$\pm$0.001 & -- & -- & authors' code crashes on this dataset \\
cmc & Acc. & 0.531$\pm$0.015 & 0.515$\pm$0.015 & -0.016 &  \\
communities & Acc. & 0.711$\pm$0.012 & 0.709$\pm$0.000 & -0.002 & one fold only (context overflow, see main text) \\
covtype & Acc. & 0.882$\pm$0.003 & -- & -- & data not released \\
credit-g & Acc. & 0.766$\pm$0.015 & 0.758$\pm$0.017 & -0.008 &  \\
eucalyptus & Acc. & 0.668$\pm$0.027 & 0.663$\pm$0.028 & -0.005 &  \\
heart & Acc. & 0.866$\pm$0.021 & 0.862$\pm$0.020 & -0.004 &  \\
junglechess & Acc. & 0.969$\pm$0.004 & 0.980$\pm$0.004 & +0.011 &  \\
myocardial & Acc. & 0.789$\pm$0.023 & 0.790$\pm$0.027 & +0.001 &  \\
pc1 & Acc. & 0.935$\pm$0.006 & 0.940$\pm$0.017 & +0.005 &  \\
tic-tac-toe & Acc. & 0.998$\pm$0.005 & 0.998$\pm$0.005 & -0.000 &  \\
vehicle & Acc. & 0.761$\pm$0.027 & 0.759$\pm$0.021 & -0.002 &  \\
airfoil\_self\_noise & N-RMSE & 0.011$\pm$0.001 & -- & -- & data not released \\
bike & N-RMSE & 0.207$\pm$0.006 & 0.224$\pm$0.037 & +0.017 &  \\
cpu\_small & N-RMSE & 0.033$\pm$0.003 & -- & -- & data not released \\
crab & N-RMSE & 0.223$\pm$0.013 & 0.222$\pm$0.014 & -0.001 &  \\
diamonds & N-RMSE & 0.134$\pm$0.002 & -- & -- & data not released \\
forest-fires & N-RMSE & 1.417$\pm$0.083 & 4.907$\pm$2.551 & +3.490 & not reproduced (48\% zero targets); excluded from the paper \\
housing & N-RMSE & 0.218$\pm$0.009 & 0.228$\pm$0.030 & +0.010 &  \\
insurance & N-RMSE & 0.381$\pm$0.028 & 0.384$\pm$0.028 & +0.003 &  \\
plasma\_retinol & N-RMSE & 0.388$\pm$0.033 & -- & -- & data not released \\
wine & N-RMSE & 0.105$\pm$0.001 & 0.105$\pm$0.001 & +0.000 &  \\
\midrule
\multicolumn{6}{l}{21 datasets re-run: median $|$diff.$|$ = 0.004, max = 0.017, 14 of 21 within $0.01$ (forest-fires excluded)} \\
\bottomrule
\end{tabular}}
\end{table}
\FloatBarrier
\section{Harness Details}
\label{sec:harness}
\llmfe{} runs from the authors' public code with three changes. We use OpenRouter instead of the OpenAI endpoint, increase the completion limit from the default 512 tokens to the limits of Table~\ref{tab:hyper} of the main text, and add to its extractor a repair already used by \shapevo: if a reply is cut off before the \texttt{return} statement, the last assigned variable is returned instead of discarding the reply. \shapevo's extractor also removes incomplete final statements and ignores attempts to \texttt{drop} nonexistent columns, a common GPT-3.5 error. The official scorer uses the same source rewrites as the search evaluator: redirecting \texttt{df\_input} reads to \texttt{df\_output} after its creation and adding \texttt{errors="ignore"} to \texttt{.drop()} calls. The Haiku runs use an updated \shapevo{} extractor that preserves indented helper functions and comments, retains complete functions in truncated replies, and fixes the input-column rewrite. The scorer applies this last fix to all backbones. It affects 33 of the 21,831 GPT-3.5 programs and 50 of the 21,849 no-SHAP programs (22,000 evaluations each, less the calls that returned no program), one DeepSeek program and none for Haiku. All affected programs failed during search, so the reported results remain unchanged. We exclude one OpenRouter provider because its coding style is incompatible with both program extractors.

The \llmfe{} paper does not state its N-RMSE normaliser. We recovered $\mathrm{mean}(y)$ by reverse-solving its published baseline column, which 5 of 6 regression datasets match to within $0.003$. \texttt{forest-fires} is the exception, and its published \llmfe{} score could not be reproduced either (Table~\ref{tab:repro}), because 48\% of its targets are zero and a few extreme values dominate the RMSE under
any split. For this reason it is left out of every table, figure and count in the paper, and only its published and re-run scores are kept in Table~\ref{tab:repro}. One detail of the evaluator affects fairness. It reads data with nullable pandas types, whereas both searches see missing values filled with zero, and programs that crash on \texttt{pd.NA} are silently dropped. We re-try such programs on zero-filled input, which changes nothing for programs that already ran and recovers folds that were discarded. For GPT-3.5 this turned one apparent \llmfe{} win, \texttt{housing}, into a tie.

The first two backbones, including both methods, the control, a pilot and three aborted \shapevo{} runs, cost about 250 US dollars in API credits. The Haiku runs, including both methods and the control, cost another 85 US dollars. Tokens and cost per method are in Sect.~\ref{sec:cost}.
\FloatBarrier
\section{Datasets}
\label{sec:datasets_supp}
The table below lists the 22 datasets with their number of rows, number of input columns and task type. The 21 public ones are the datasets of the \llmfe{} benchmark for which data and task descriptions were released, without \texttt{forest-fires} (Sect.~\ref{sec:harness}).
\begin{table}[h]
\centering
\caption{Datasets, with $n$ rows and $p$ input columns. \textsc{synthetic} is a dataset generated for this study, unseen by any LLM during pretraining and not publicly available. All others are the public LLM-FE benchmark tasks.}
\label{tab:datasets}
\scriptsize
\begin{tabular}{lrrl}
\toprule
Dataset & $n$ & $p$ & Task \\
\midrule
adult & 48,842 & 14 & classification \\
balance-scale & 625 & 4 & classification \\
bank & 45,211 & 16 & classification \\
bike & 17,379 & 12 & regression \\
blood & 748 & 4 & classification \\
breast-w & 699 & 9 & classification \\
car & 1,728 & 6 & classification \\
cmc & 1,473 & 9 & classification \\
communities & 1,994 & 102 & classification \\
crab & 3,893 & 8 & regression \\
credit-g & 1,000 & 20 & classification \\
eucalyptus & 736 & 19 & classification \\
heart & 918 & 11 & classification \\
housing & 20,640 & 9 & regression \\
insurance & 1,338 & 6 & regression \\
junglechess & 44,819 & 6 & classification \\
myocardial & 686 & 91 & classification \\
pc1 & 1,109 & 21 & classification \\
synthetic (grouped CV) & 21,208 & 19 & classification \\
tic-tac-toe & 958 & 9 & classification \\
vehicle & 846 & 18 & classification \\
wine & 6,497 & 12 & regression \\
\bottomrule
\end{tabular}
\end{table}
\FloatBarrier
\section{Convergence}
\label{sec:convergence_supp}
This section shows the search-speed comparison of Sect.~\ref{sec:anytime} of the main paper in more detail: the speed-up at equal level along the whole budget, how the LLM tokens of Table~\ref{tab:summary} are counted, and for every dataset the score of every evaluated program against the number of evaluations, the best-so-far score against wall-clock time, and the wall-clock time each method needs to reach 95\% of its own final gain.
\begin{figure}[tbp]
\centering
\includegraphics[width=\textwidth]{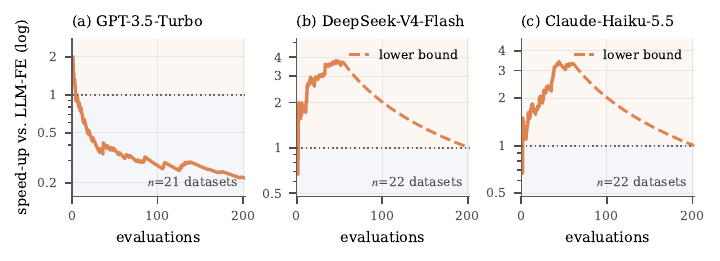}
\caption{Speed-up at equal level, from the curves of Fig.~\ref{fig:convergence} of the main text: for the level \shapevo{} has after a given number of programs, how many times more programs \llmfe{} needs to reach the same level. Values above 1 mean \shapevo{} gets there with fewer programs. Where the curve is dashed, SHAPEvo-FE is above the final level of LLM-FE, which never reaches it within its budget of 200. The plotted value is then $200/n$, a lower bound on the true speed-up, which is why the dashed part falls to 1 at $n=200$.}
\label{fig:speedup}
\end{figure}

Table~\ref{tab:summary} of the main text counts LLM tokens with one tokenizer (tiktoken, \texttt{cl100k\_base}) on the prompt and reply text of every call, for both methods, because the \llmfe{} code records no token usage. For \shapevo{} the saved prompts and replies are counted directly. For \llmfe{} the prompt is reconstructed as its code builds it, from the role and instruction block, the task sentence, one description line per column, ten example rows, the two sampled parent programs (taken as the mean program length of that fold) and the header of the function to complete, and the reply is the stored program. The counts approximate what the provider bills (the providers' own tokenizers split the same text differently: on \texttt{heart}, fold 1, our count divided by the prompt tokens the provider billed is 1.00 for GPT-3.5, 0.92 for DeepSeek and 0.64 for Haiku), but the same tokenizer is used for both methods of a backbone, so the ratio is fair. Wall-clock minutes in Table~3 come from the same runs, which for a given backbone were executed through the same gateway in the same week.

\begin{figure}[p]
\centering
\includegraphics[width=\textwidth]{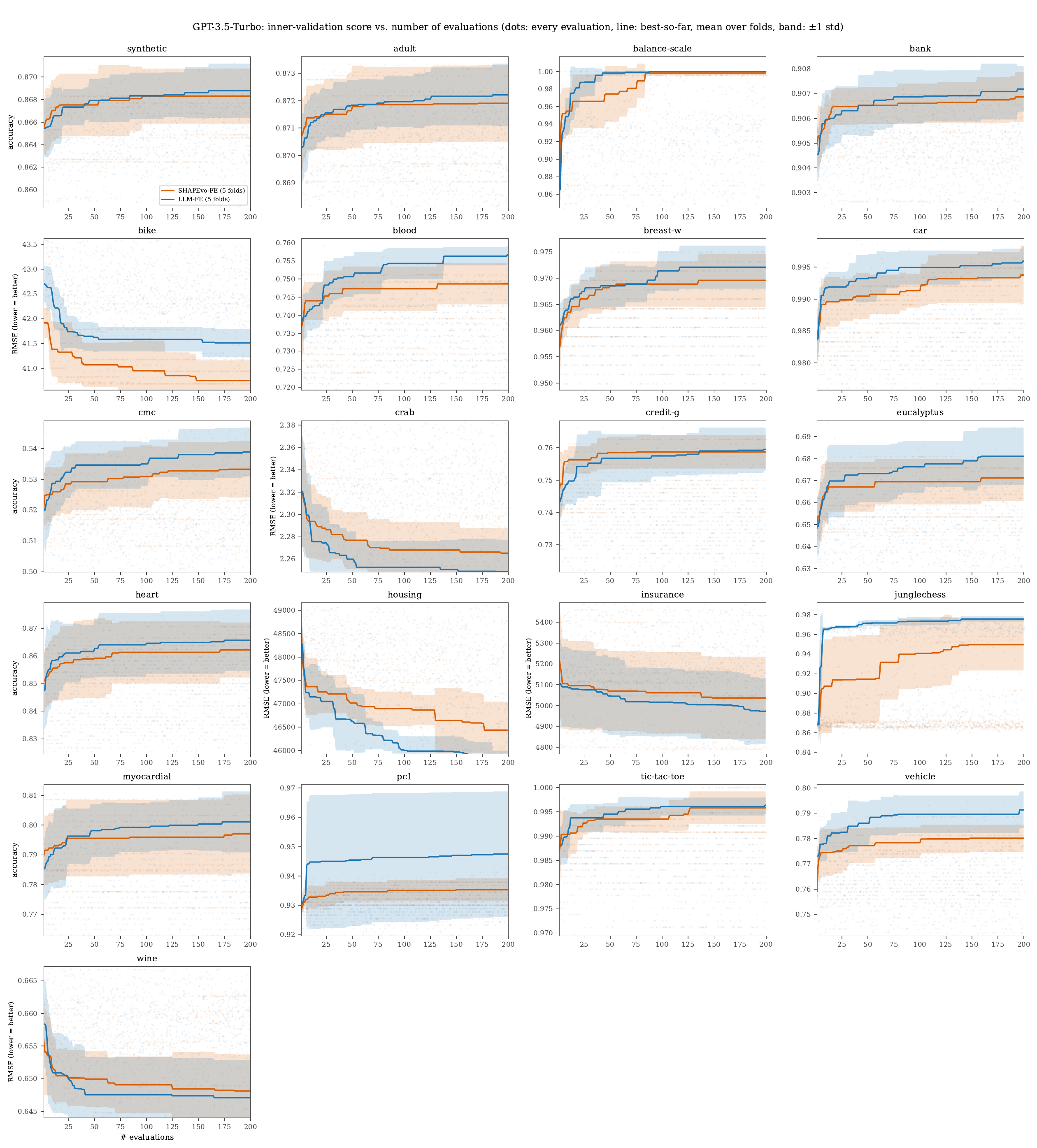}
\caption{Score against the number of evaluations for every dataset with the GPT-3.5-Turbo backbone. Each dot is the inner-CV score of one evaluated program, pooled over the five outer folds. The lines are the best-so-far score, mean over folds, with a band of one standard deviation. Regression datasets are drawn as RMSE, lower is better. The y-axis covers the lines and the central 90\% of the dots, so a few programs with extreme scores fall outside the panel. \texttt{communities} is omitted because \llmfe{} completed one fold.}
\end{figure}
\begin{figure}[p]
\centering
\includegraphics[width=\textwidth]{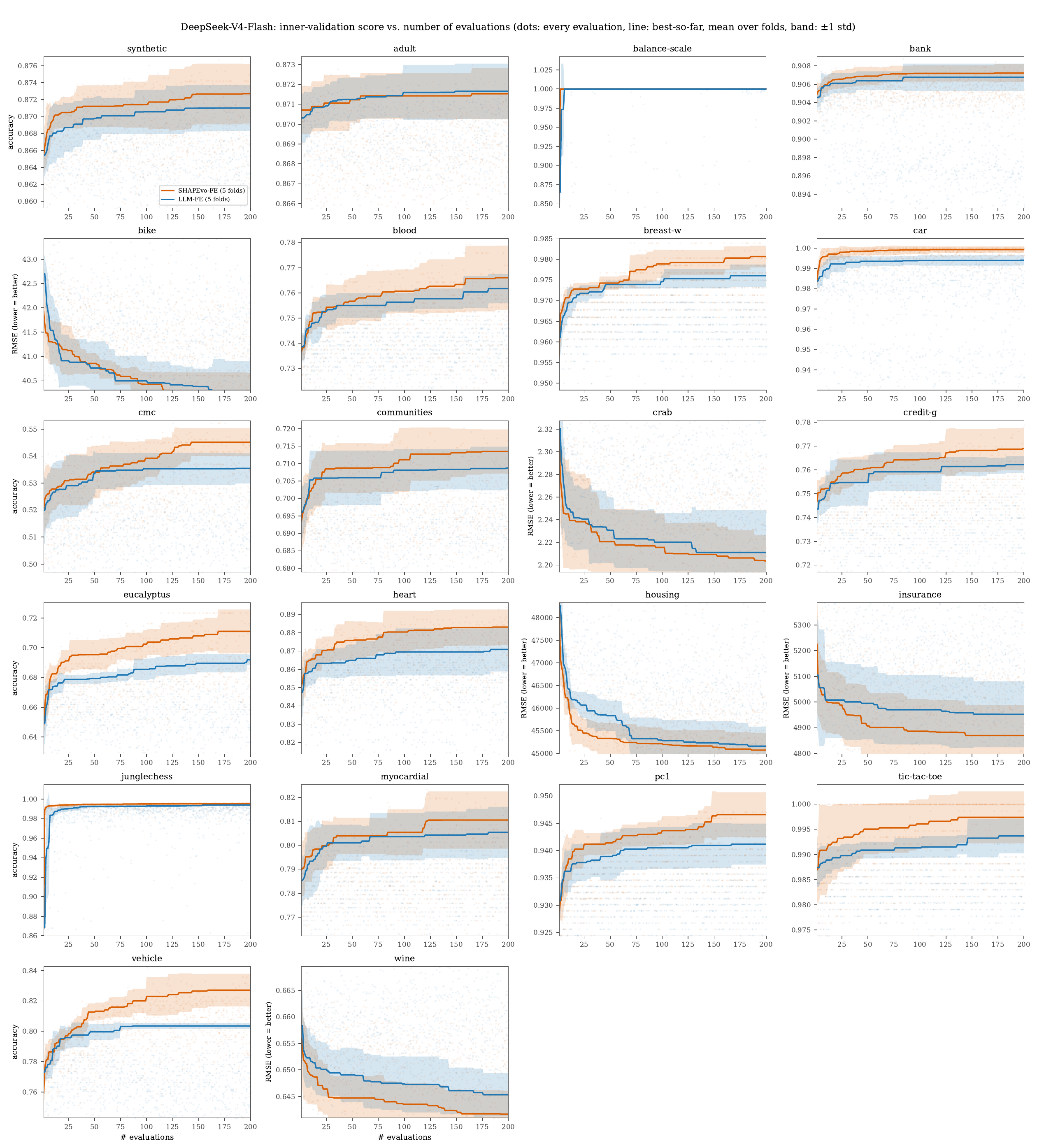}
\caption{As the previous figure, DeepSeek-V4-Flash backbone.}
\end{figure}
\begin{figure}[p]
\centering
\includegraphics[width=\textwidth]{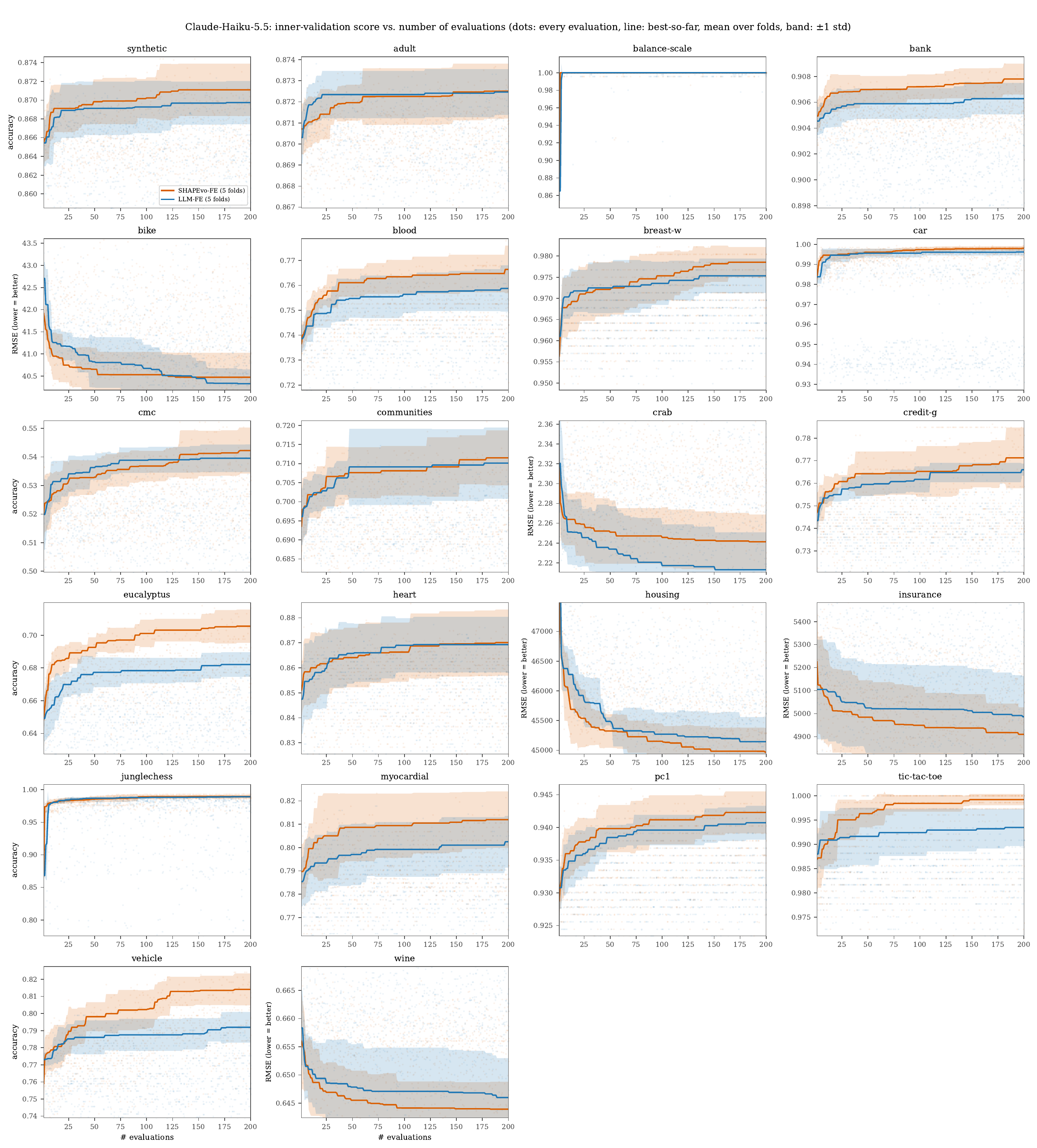}
\caption{As the previous figure, Claude-Haiku-5.5 backbone.}
\end{figure}
\begin{figure}[p]
\centering
\includegraphics[width=\textwidth]{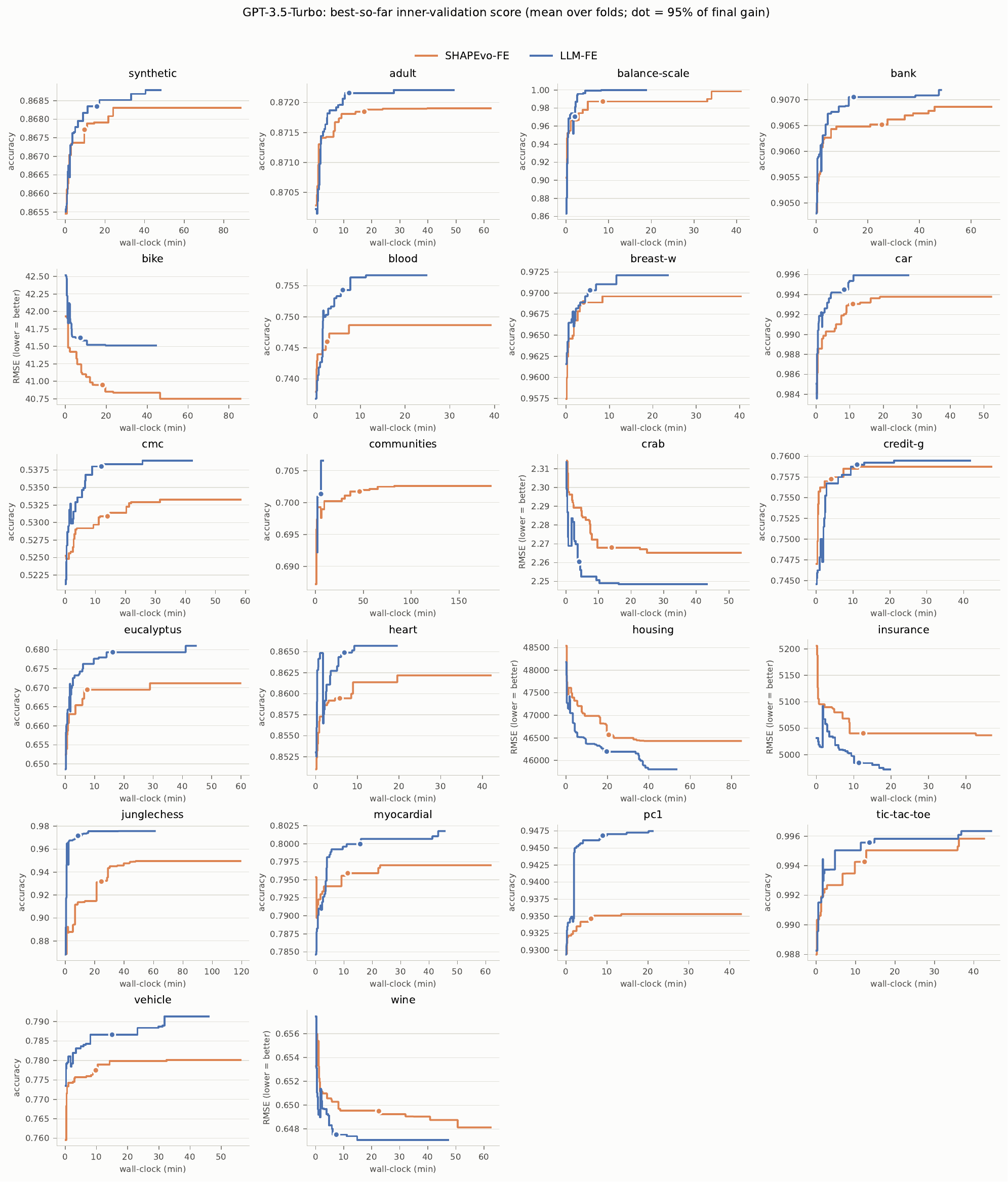}
\caption{Best-so-far inner-CV score against wall-clock time for every dataset with the GPT-3.5-Turbo backbone, mean over the five outer folds. The dot marks the time at which a method reaches 95\% of its own final gain, the $t_{95}$ of Table~\ref{tab:ttc}. Both methods were run through the same API gateway in the same week.}
\end{figure}
\begin{figure}[p]
\centering
\includegraphics[width=\textwidth]{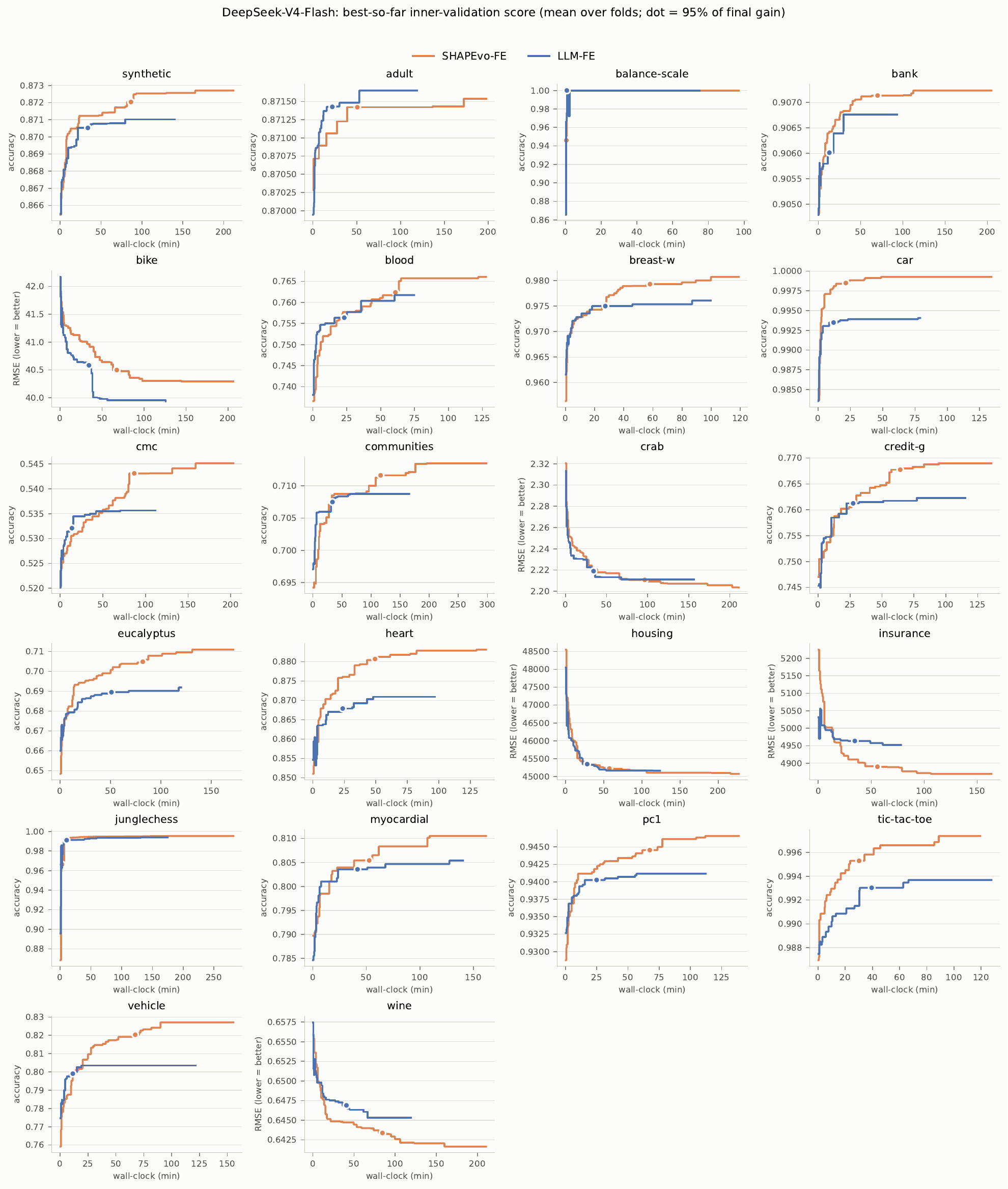}
\caption{As the previous figure, DeepSeek-V4-Flash backbone.}
\end{figure}
\begin{figure}[p]
\centering
\includegraphics[width=\textwidth]{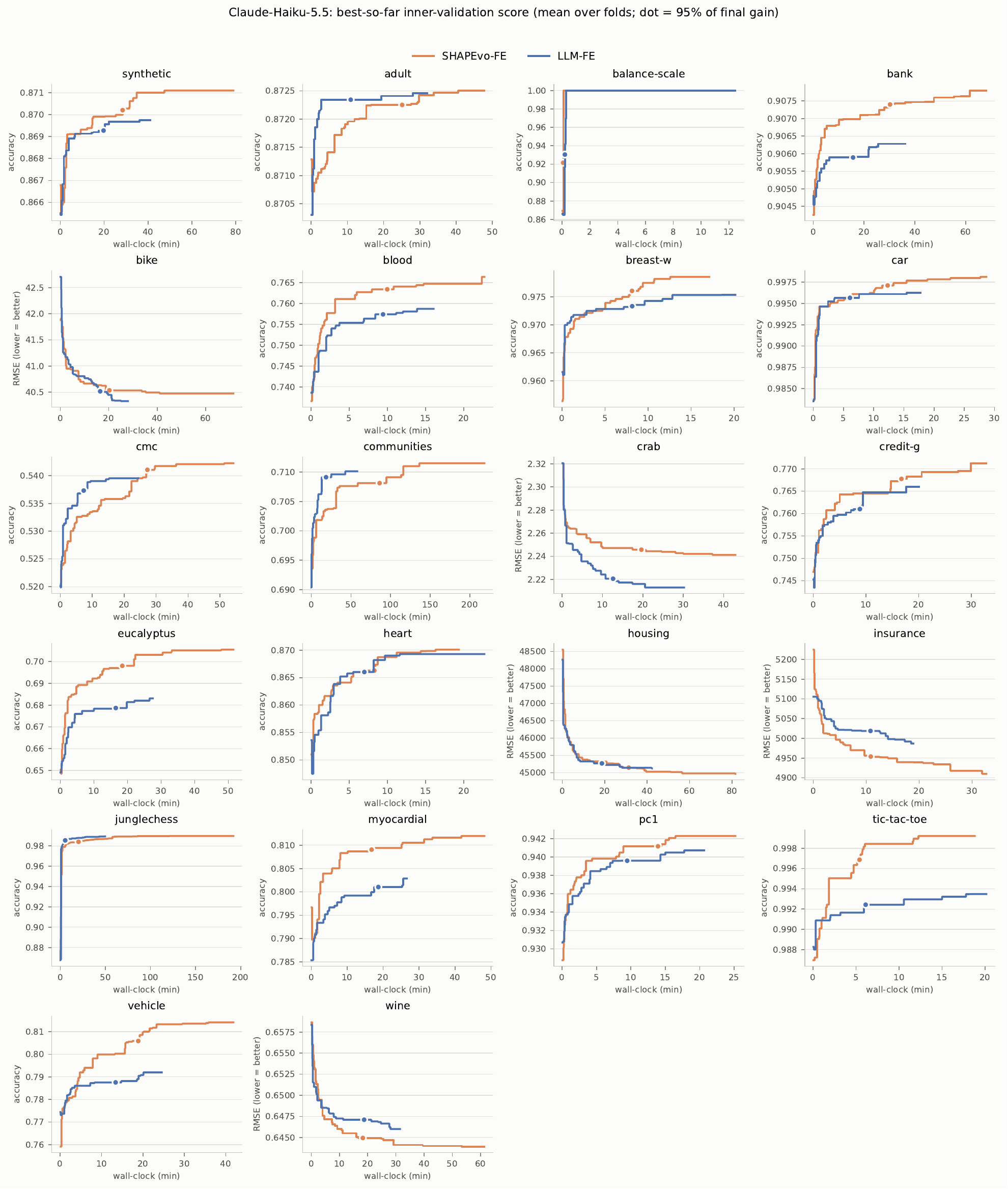}
\caption{As the previous figure, Claude-Haiku-5.5 backbone.}
\end{figure}
\begin{table}[p]
\centering
\caption{Wall-clock time in minutes to reach 95\% of a method's own final gain ($t_{95}$) and total run time per fold, median over folds, per backbone. Both methods were run through the same API gateway in the same week. \shapevo's longer run times with DeepSeek come from its longer replies, a median of 46 features per program, and from the SHAP computation after every evaluation.}
\label{tab:ttc}
\scriptsize
\resizebox{\textwidth}{!}{%
\begin{tabular}{@{}c@{}}\textbf{GPT-3.5-Turbo}\\[3pt]\begin{tabular}{lrrrr}
\toprule
 & \multicolumn{2}{c}{LLM-FE} & \multicolumn{2}{c}{SHAPEvo-FE} \\
Dataset & $t_{95}$ (min) & total (min) & $t_{95}$ (min) & total (min) \\
\midrule
adult & 9.6 & 20.9 & 14.2 & 27.4 \\
balance-scale & 2.2 & 14.0 & 3.0 & 15.3 \\
bank & 8.7 & 18.9 & 34.2 & 35.7 \\
bike & 3.1 & 16.6 & 11.9 & 32.7 \\
blood & 5.6 & 11.5 & 2.2 & 12.0 \\
breast-w & 5.0 & 12.4 & 3.2 & 15.0 \\
car & 9.5 & 12.0 & 9.6 & 16.3 \\
cmc & 8.8 & 13.8 & 11.0 & 24.4 \\
communities & 5.7 & 7.8 & 47.1 & 86.0 \\
crab & 4.4 & 17.0 & 7.9 & 23.5 \\
credit-g & 10.3 & 15.6 & 3.7 & 23.3 \\
eucalyptus & 11.4 & 18.5 & 0.9 & 28.2 \\
heart & 6.9 & 12.5 & 2.7 & 17.6 \\
housing & 16.6 & 20.8 & 19.5 & 32.9 \\
insurance & 10.0 & 11.4 & 6.9 & 16.6 \\
junglechess & 7.5 & 30.1 & 30.1 & 61.2 \\
myocardial & 10.9 & 21.9 & 9.0 & 28.3 \\
pc1 & 8.6 & 13.1 & 5.5 & 17.8 \\
synthetic & 11.1 & 21.5 & 9.3 & 37.0 \\
tic-tac-toe & 11.3 & 19.2 & 9.7 & 16.0 \\
vehicle & 8.0 & 21.0 & 3.0 & 22.7 \\
wine & 6.0 & 19.8 & 22.7 & 27.3 \\
\midrule
median & 8.6 & & 9.1 & \\
\bottomrule
\end{tabular}\end{tabular}\hspace{1em}%
\begin{tabular}{@{}c@{}}\textbf{DeepSeek-V4-Flash}\\[3pt]\begin{tabular}{lrrrr}
\toprule
 & \multicolumn{2}{c}{LLM-FE} & \multicolumn{2}{c}{SHAPEvo-FE} \\
Dataset & $t_{95}$ (min) & total (min) & $t_{95}$ (min) & total (min) \\
\midrule
adult & 11.8 & 48.4 & 27.6 & 130.6 \\
balance-scale & 0.4 & 38.7 & 0.4 & 75.9 \\
bank & 11.0 & 41.3 & 50.3 & 132.6 \\
bike & 38.3 & 51.4 & 81.7 & 143.3 \\
blood & 15.9 & 46.7 & 63.1 & 83.2 \\
breast-w & 16.6 & 42.0 & 37.5 & 78.9 \\
car & 15.1 & 50.1 & 21.4 & 91.9 \\
cmc & 8.0 & 45.6 & 79.5 & 126.4 \\
communities & 35.2 & 75.6 & 108.1 & 224.9 \\
crab & 31.5 & 60.4 & 77.0 & 131.3 \\
credit-g & 16.2 & 39.6 & 57.3 & 94.8 \\
eucalyptus & 30.1 & 47.5 & 82.6 & 107.8 \\
heart & 23.2 & 44.8 & 50.8 & 87.9 \\
housing & 22.1 & 54.5 & 54.6 & 145.0 \\
insurance & 28.0 & 46.7 & 55.3 & 101.8 \\
junglechess & 3.6 & 89.4 & 1.5 & 213.1 \\
myocardial & 22.3 & 63.9 & 55.2 & 101.0 \\
pc1 & 15.3 & 40.5 & 61.3 & 89.2 \\
synthetic & 21.5 & 63.0 & 89.0 & 134.0 \\
tic-tac-toe & 30.4 & 53.1 & 19.8 & 84.6 \\
vehicle & 13.9 & 52.3 & 70.7 & 102.4 \\
wine & 41.6 & 49.9 & 94.5 & 126.5 \\
\midrule
median & 19.0 & & 56.3 & \\
\bottomrule
\end{tabular}\end{tabular}\hspace{1em}%
\begin{tabular}{@{}c@{}}\textbf{Claude-Haiku-5.5}\\[3pt]\begin{tabular}{lrrrr}
\toprule
 & \multicolumn{2}{c}{LLM-FE} & \multicolumn{2}{c}{SHAPEvo-FE} \\
Dataset & $t_{95}$ (min) & total (min) & $t_{95}$ (min) & total (min) \\
\midrule
adult & 2.7 & 31.4 & 29.6 & 45.7 \\
balance-scale & 0.2 & 12.1 & 0.1 & 8.4 \\
bank & 21.7 & 33.2 & 27.9 & 52.2 \\
bike & 19.2 & 27.9 & 13.7 & 52.4 \\
blood & 8.4 & 15.1 & 10.8 & 20.8 \\
breast-w & 9.6 & 18.6 & 9.4 & 15.4 \\
car & 3.5 & 16.6 & 13.2 & 27.0 \\
cmc & 6.4 & 19.6 & 26.8 & 41.9 \\
communities & 12.9 & 50.1 & 113.1 & 150.7 \\
crab & 13.1 & 27.9 & 17.4 & 40.5 \\
credit-g & 9.2 & 18.5 & 17.7 & 22.7 \\
eucalyptus & 19.8 & 26.3 & 21.9 & 48.1 \\
heart & 8.2 & 21.4 & 8.4 & 17.4 \\
housing & 19.8 & 35.2 & 38.5 & 63.2 \\
insurance & 14.1 & 17.3 & 6.0 & 28.9 \\
junglechess & 4.5 & 48.1 & 10.4 & 126.5 \\
myocardial & 17.5 & 26.3 & 5.7 & 41.6 \\
pc1 & 7.4 & 20.3 & 15.0 & 20.5 \\
synthetic & 20.2 & 38.9 & 31.6 & 59.4 \\
tic-tac-toe & 2.0 & 19.9 & 5.5 & 17.3 \\
vehicle & 18.9 & 23.6 & 19.1 & 38.0 \\
wine & 24.7 & 30.5 & 16.7 & 53.4 \\
\midrule
median & 11.3 & & 15.9 & \\
\bottomrule
\end{tabular}\end{tabular}}
\end{table}
\FloatBarrier
\section{Held-Out Scores of the No-SHAP Control}
Tables~\ref{tab:noshap_gpt35} to~\ref{tab:noshap_haiku} give the held-out scores of the no-SHAP control of Sect.~\ref{sec:protocol} of the main paper next to those of \llmfe{} and the full \shapevo, per dataset and backbone, with the same evaluator, splits and ensemble as Table~\ref{tab:h2h}. The last two columns are the differences between the full method and the control and between the control and \llmfe, signed so that a positive value favours the first of the two. With GPT-3.5 the control and the full method are on par, with DeepSeek and Haiku the full method is ahead of the control on more datasets than it trails, and on every backbone the control alone is on par with \llmfe. The \llmfe{} wins of the control on \texttt{car} and \texttt{bike} with DeepSeek and on \texttt{car}, \texttt{bike} and \texttt{tic-tac-toe} with Haiku are the collapsed \llmfe{} folds discussed in Sect.~\ref{sec:h2h} of the main paper.
\begin{table}[p]
\centering
\caption{No-SHAP control with GPT-3.5-Turbo, held-out top-3 ensemble scores (mean $\pm$ std over folds), \texttt{communities} excluded as in the main paper. Bold marks $|\Delta|>0.01$.}
\label{tab:noshap_gpt35}
\scriptsize
\resizebox{\textwidth}{!}{\begin{tabular}{llrrrrr}
\toprule
Dataset & Metric & LLM-FE & SHAPEvo-FE & SHAPEvo-FE (no SHAP) & $\Delta$(full$-$noSHAP) & $\Delta$(noSHAP$-$LLM-FE) \\
\midrule
adult & Acc.\ $\uparrow$ & 0.872$\pm$0.004 & 0.872$\pm$0.002 & 0.874$\pm$0.002 & -0.002 & +0.002 \\
balance-scale & Acc.\ $\uparrow$ & 1.000$\pm$0.000 & 1.000$\pm$0.000 & 1.000$\pm$0.000 & +0.000 & +0.000 \\
bank & Acc.\ $\uparrow$ & 0.907$\pm$0.003 & 0.906$\pm$0.003 & 0.906$\pm$0.002 & -0.000 & -0.001 \\
blood & Acc.\ $\uparrow$ & 0.741$\pm$0.030 & 0.735$\pm$0.029 & 0.745$\pm$0.028 & -0.009 & +0.004 \\
breast-w & Acc.\ $\uparrow$ & 0.957$\pm$0.010 & 0.959$\pm$0.003 & 0.963$\pm$0.009 & -0.004 & +0.006 \\
car & Acc.\ $\uparrow$ & 1.000$\pm$0.000 & 0.999$\pm$0.003 & 0.998$\pm$0.003 & +0.001 & -0.002 \\
cmc & Acc.\ $\uparrow$ & 0.515$\pm$0.015 & 0.521$\pm$0.010 & 0.521$\pm$0.019 & -0.000 & +0.006 \\
credit-g & Acc.\ $\uparrow$ & 0.758$\pm$0.017 & 0.749$\pm$0.018 & 0.762$\pm$0.039 & \textbf{-0.013} & +0.004 \\
eucalyptus & Acc.\ $\uparrow$ & 0.663$\pm$0.028 & 0.673$\pm$0.029 & 0.675$\pm$0.031 & -0.003 & \textbf{+0.012} \\
heart & Acc.\ $\uparrow$ & 0.862$\pm$0.020 & 0.857$\pm$0.023 & 0.856$\pm$0.024 & +0.001 & -0.005 \\
junglechess & Acc.\ $\uparrow$ & 0.980$\pm$0.004 & 0.954$\pm$0.026 & 0.972$\pm$0.005 & \textbf{-0.018} & -0.008 \\
myocardial & Acc.\ $\uparrow$ & 0.790$\pm$0.027 & 0.776$\pm$0.031 & 0.787$\pm$0.024 & \textbf{-0.012} & -0.003 \\
pc1 & Acc.\ $\uparrow$ & 0.940$\pm$0.017 & 0.931$\pm$0.005 & 0.930$\pm$0.009 & +0.002 & \textbf{-0.011} \\
synthetic & Acc.\ $\uparrow$ & 0.872$\pm$0.003 & 0.865$\pm$0.007 & 0.869$\pm$0.004 & -0.004 & -0.002 \\
tic-tac-toe & Acc.\ $\uparrow$ & 0.998$\pm$0.005 & 0.990$\pm$0.010 & 0.997$\pm$0.005 & -0.007 & -0.001 \\
vehicle & Acc.\ $\uparrow$ & 0.759$\pm$0.021 & 0.766$\pm$0.030 & 0.759$\pm$0.026 & +0.007 & -0.000 \\
bike & N-RMSE $\downarrow$ & 0.224$\pm$0.037 & 0.211$\pm$0.006 & 0.213$\pm$0.007 & +0.001 & \textbf{+0.012} \\
crab & N-RMSE $\downarrow$ & 0.222$\pm$0.014 & 0.232$\pm$0.015 & 0.231$\pm$0.014 & -0.001 & -0.009 \\
housing & N-RMSE $\downarrow$ & 0.228$\pm$0.030 & 0.218$\pm$0.008 & 0.228$\pm$0.010 & \textbf{+0.010} & +0.000 \\
insurance & N-RMSE $\downarrow$ & 0.384$\pm$0.028 & 0.385$\pm$0.023 & 0.388$\pm$0.024 & +0.003 & -0.004 \\
wine & N-RMSE $\downarrow$ & 0.105$\pm$0.001 & 0.108$\pm$0.001 & 0.109$\pm$0.001 & +0.001 & -0.004 \\
\midrule
\multicolumn{7}{l}{full vs.\ no-SHAP: full better 1 / no-SHAP better 3 / tie 17; no-SHAP vs.\ LLM-FE: no-SHAP better 2 / LLM-FE better 1 / tie 18 ($|\Delta|>0.01$)} \\
\bottomrule
\end{tabular}}
\end{table}
\begin{table}[p]
\centering
\caption{As the previous table, DeepSeek-V4-Flash backbone.}
\label{tab:noshap_deepseek}
\scriptsize
\resizebox{\textwidth}{!}{\begin{tabular}{llrrrrr}
\toprule
Dataset & Metric & LLM-FE & SHAPEvo-FE & SHAPEvo-FE (no SHAP) & $\Delta$(full$-$noSHAP) & $\Delta$(noSHAP$-$LLM-FE) \\
\midrule
adult & Acc.\ $\uparrow$ & 0.868$\pm$0.010 & 0.873$\pm$0.002 & 0.872$\pm$0.002 & +0.001 & +0.004 \\
balance-scale & Acc.\ $\uparrow$ & 1.000$\pm$0.000 & 1.000$\pm$0.000 & 1.000$\pm$0.000 & +0.000 & +0.000 \\
bank & Acc.\ $\uparrow$ & 0.907$\pm$0.004 & 0.907$\pm$0.002 & 0.905$\pm$0.003 & +0.002 & -0.002 \\
blood & Acc.\ $\uparrow$ & 0.749$\pm$0.030 & 0.755$\pm$0.026 & 0.738$\pm$0.026 & \textbf{+0.017} & \textbf{-0.011} \\
breast-w & Acc.\ $\uparrow$ & 0.971$\pm$0.011 & 0.967$\pm$0.004 & 0.964$\pm$0.009 & +0.003 & -0.007 \\
car & Acc.\ $\uparrow$ & 0.908$\pm$0.174 & 1.000$\pm$0.000 & 0.998$\pm$0.003 & +0.002 & \textbf{+0.090} \\
cmc & Acc.\ $\uparrow$ & 0.526$\pm$0.027 & 0.511$\pm$0.017 & 0.534$\pm$0.014 & \textbf{-0.023} & +0.008 \\
communities & Acc.\ $\uparrow$ & 0.706$\pm$0.016 & 0.694$\pm$0.015 & 0.700$\pm$0.015 & -0.006 & -0.006 \\
credit-g & Acc.\ $\uparrow$ & 0.758$\pm$0.019 & 0.740$\pm$0.010 & 0.752$\pm$0.016 & \textbf{-0.012} & -0.006 \\
eucalyptus & Acc.\ $\uparrow$ & 0.667$\pm$0.034 & 0.674$\pm$0.018 & 0.644$\pm$0.021 & \textbf{+0.030} & \textbf{-0.023} \\
heart & Acc.\ $\uparrow$ & 0.852$\pm$0.024 & 0.854$\pm$0.031 & 0.854$\pm$0.029 & +0.000 & +0.002 \\
junglechess & Acc.\ $\uparrow$ & 0.997$\pm$0.001 & 0.997$\pm$0.001 & 0.997$\pm$0.001 & +0.000 & +0.000 \\
myocardial & Acc.\ $\uparrow$ & 0.796$\pm$0.019 & 0.789$\pm$0.015 & 0.783$\pm$0.031 & +0.006 & \textbf{-0.013} \\
pc1 & Acc.\ $\uparrow$ & 0.931$\pm$0.002 & 0.938$\pm$0.010 & 0.936$\pm$0.007 & +0.002 & +0.005 \\
synthetic & Acc.\ $\uparrow$ & 0.867$\pm$0.007 & 0.871$\pm$0.005 & 0.867$\pm$0.004 & +0.004 & +0.000 \\
tic-tac-toe & Acc.\ $\uparrow$ & 0.987$\pm$0.011 & 0.998$\pm$0.005 & 0.996$\pm$0.007 & +0.002 & +0.008 \\
vehicle & Acc.\ $\uparrow$ & 0.775$\pm$0.018 & 0.801$\pm$0.033 & 0.786$\pm$0.020 & \textbf{+0.015} & \textbf{+0.011} \\
bike & N-RMSE $\downarrow$ & 0.297$\pm$0.116 & 0.204$\pm$0.004 & 0.204$\pm$0.005 & +0.001 & \textbf{+0.093} \\
crab & N-RMSE $\downarrow$ & 0.218$\pm$0.015 & 0.224$\pm$0.013 & 0.228$\pm$0.015 & +0.004 & \textbf{-0.010} \\
housing & N-RMSE $\downarrow$ & 0.245$\pm$0.017 & 0.211$\pm$0.005 & 0.222$\pm$0.018 & \textbf{+0.011} & \textbf{+0.023} \\
insurance & N-RMSE $\downarrow$ & 0.396$\pm$0.047 & 0.380$\pm$0.034 & 0.384$\pm$0.014 & +0.004 & \textbf{+0.012} \\
wine & N-RMSE $\downarrow$ & 0.105$\pm$0.001 & 0.106$\pm$0.003 & 0.108$\pm$0.001 & +0.002 & -0.003 \\
\midrule
\multicolumn{7}{l}{full vs.\ no-SHAP: full better 4 / no-SHAP better 2 / tie 16; no-SHAP vs.\ LLM-FE: no-SHAP better 5 / LLM-FE better 4 / tie 13 ($|\Delta|>0.01$)} \\
\bottomrule
\end{tabular}}
\end{table}
\begin{table}[p]
\centering
\caption{As the previous table, Claude-Haiku-5.5 backbone.}
\label{tab:noshap_haiku}
\scriptsize
\resizebox{\textwidth}{!}{\begin{tabular}{llrrrrr}
\toprule
Dataset & Metric & LLM-FE & SHAPEvo-FE & SHAPEvo-FE (no SHAP) & $\Delta$(full$-$noSHAP) & $\Delta$(noSHAP$-$LLM-FE) \\
\midrule
adult & Acc.\ $\uparrow$ & 0.874$\pm$0.003 & 0.874$\pm$0.003 & 0.874$\pm$0.003 & -0.000 & +0.000 \\
balance-scale & Acc.\ $\uparrow$ & 1.000$\pm$0.000 & 1.000$\pm$0.000 & 1.000$\pm$0.000 & +0.000 & +0.000 \\
bank & Acc.\ $\uparrow$ & 0.894$\pm$0.019 & 0.909$\pm$0.005 & 0.907$\pm$0.003 & +0.002 & \textbf{+0.013} \\
blood & Acc.\ $\uparrow$ & 0.751$\pm$0.020 & 0.759$\pm$0.033 & 0.747$\pm$0.026 & \textbf{+0.012} & -0.004 \\
breast-w & Acc.\ $\uparrow$ & 0.966$\pm$0.009 & 0.961$\pm$0.012 & 0.966$\pm$0.012 & -0.004 & -0.000 \\
car & Acc.\ $\uparrow$ & 0.957$\pm$0.093 & 0.998$\pm$0.004 & 0.998$\pm$0.003 & -0.000 & \textbf{+0.041} \\
cmc & Acc.\ $\uparrow$ & 0.521$\pm$0.022 & 0.495$\pm$0.021 & 0.531$\pm$0.016 & \textbf{-0.036} & +0.009 \\
communities & Acc.\ $\uparrow$ & 0.706$\pm$0.022 & 0.708$\pm$0.022 & 0.703$\pm$0.018 & +0.005 & -0.003 \\
credit-g & Acc.\ $\uparrow$ & 0.741$\pm$0.021 & 0.740$\pm$0.017 & 0.748$\pm$0.021 & -0.008 & +0.007 \\
eucalyptus & Acc.\ $\uparrow$ & 0.671$\pm$0.039 & 0.679$\pm$0.021 & 0.659$\pm$0.038 & \textbf{+0.020} & \textbf{-0.012} \\
heart & Acc.\ $\uparrow$ & 0.855$\pm$0.023 & 0.861$\pm$0.026 & 0.846$\pm$0.014 & \textbf{+0.014} & -0.009 \\
junglechess & Acc.\ $\uparrow$ & 0.992$\pm$0.002 & 0.993$\pm$0.001 & 0.995$\pm$0.002 & -0.001 & +0.002 \\
myocardial & Acc.\ $\uparrow$ & 0.780$\pm$0.033 & 0.778$\pm$0.026 & 0.794$\pm$0.021 & \textbf{-0.016} & \textbf{+0.015} \\
pc1 & Acc.\ $\uparrow$ & 0.932$\pm$0.013 & 0.931$\pm$0.017 & 0.935$\pm$0.007 & -0.005 & +0.003 \\
synthetic & Acc.\ $\uparrow$ & 0.868$\pm$0.008 & 0.871$\pm$0.004 & 0.867$\pm$0.008 & +0.004 & -0.001 \\
tic-tac-toe & Acc.\ $\uparrow$ & 0.903$\pm$0.194 & 0.999$\pm$0.002 & 0.994$\pm$0.011 & +0.005 & \textbf{+0.091} \\
vehicle & Acc.\ $\uparrow$ & 0.772$\pm$0.026 & 0.790$\pm$0.017 & 0.775$\pm$0.039 & \textbf{+0.014} & +0.004 \\
bike & N-RMSE $\downarrow$ & 0.244$\pm$0.061 & 0.202$\pm$0.006 & 0.208$\pm$0.006 & +0.006 & \textbf{+0.036} \\
crab & N-RMSE $\downarrow$ & 0.220$\pm$0.013 & 0.220$\pm$0.015 & 0.228$\pm$0.010 & +0.007 & -0.007 \\
housing & N-RMSE $\downarrow$ & 0.235$\pm$0.024 & 0.214$\pm$0.011 & 0.225$\pm$0.017 & \textbf{+0.011} & \textbf{+0.010} \\
insurance & N-RMSE $\downarrow$ & 0.387$\pm$0.030 & 0.381$\pm$0.021 & 0.385$\pm$0.021 & +0.004 & +0.002 \\
wine & N-RMSE $\downarrow$ & 0.105$\pm$0.002 & 0.106$\pm$0.002 & 0.107$\pm$0.002 & +0.001 & -0.002 \\
\midrule
\multicolumn{7}{l}{full vs.\ no-SHAP: full better 5 / no-SHAP better 2 / tie 15; no-SHAP vs.\ LLM-FE: no-SHAP better 6 / LLM-FE better 1 / tie 15 ($|\Delta|>0.01$)} \\
\bottomrule
\end{tabular}}
\end{table}

\FloatBarrier
\section{Post-hoc Pruning of the Best Programs}
With DeepSeek the best \shapevo{} program of a fold adds a median of 46 features and the best \llmfe{} program 18 (Table~\ref{tab:features} of the main paper). To see how many of them the model needs, we took the best program of every fold, applied it to the training and held-out rows of that fold, trained one XGBoost model with seed 42 and scored it on the held-out rows, as in the official evaluation but with a single program instead of the top-3 ensemble. We then explained the model with TreeSHAP on the training rows, turned the mean absolute SHAP values into shares that sum to one, and retrained after dropping every added feature below a share threshold. Original columns are always kept. The two thresholds are the ones the SHAP report uses, 0.05 for \textsc{keep} and 0.01 for \textsc{keep} plus \textsc{refine}. A fourth variant keeps the fewest added features whose shares together cover 80\% of the attribution of all added features, a rank-based rule that does not depend on the thresholds. Table~\ref{tab:prune} gives the median number of added features and the mean change of the held-out score, averaged over folds and then over datasets. Positive values mean that the pruned model scores better. A best program that fails outside the search harness is skipped, which happened on 7 of the 315 \llmfe{} folds and on no \shapevo{} fold, besides the four GPT-3.5 \texttt{communities} folds that returned no program.

Keeping only the \textsc{keep} features leaves 1 to 4 added features per program for every backbone and method, and the held-out score moves by less than 0.01 on 15 to 20 of the 21 datasets. The largest drops are \texttt{junglechess} ($-0.039$) and \texttt{vehicle} ($-0.017$) for \shapevo{} with DeepSeek and \texttt{car} and \texttt{heart} (both $-0.024$) for \llmfe{} with DeepSeek. The largest gains are \texttt{credit-g} ($+0.033$) and \texttt{cmc} ($+0.018$) for \shapevo{} with DeepSeek. With Claude-Haiku-5.5 the picture is the same at a smaller program size. \llmfe{} adds 14 features and \shapevo{} 11, of which 3 to 4 carry a \textsc{keep} share, and the largest drop after keeping only those is again \texttt{junglechess}, $-0.068$ for \shapevo{} and $-0.029$ for \llmfe. Keeping the \textsc{keep} and \textsc{refine} features changes the score by less than 0.01 on 20 or 21 of the 21 datasets in every row. The rank-based rule keeps 2 added features with GPT-3.5, 8.6 and 17.2 with DeepSeek and 7.0 and 5.8 with Haiku for \llmfe{} and \shapevo{} (Table~\ref{tab:features} of the main text, which takes medians instead of means over folds, gives 8 and 17.5 and 7 and 5), and the score moves by less than 0.01 on 18 to 21 of the 21 datasets, the largest drop being \texttt{eucalyptus}, $-0.020$ for \llmfe{} and $-0.019$ for \shapevo{} with Haiku. Dropping all added features costs 0.008 to 0.014 on average in five of the six rows. Most of the features a DeepSeek program adds are therefore idle. The model leans on a handful of them and the rest neither help nor hurt. Neither search penalises program size, which is why the programs grow.
\begin{table}[h]
\centering
\caption{Post-hoc TreeSHAP pruning of the best program of each fold. Added features are medians over datasets of the per-dataset mean over folds. $\Delta$ is the change of the single-program held-out score after pruning, mean over 21 datasets, positive when the pruned model is better, and the count gives the datasets on which $|\Delta|\le 0.01$. The last column drops every added feature. The top 80\% variant keeps the fewest added features whose shares together cover 80\% of the attribution all added features receive.}
\label{tab:prune}
\scriptsize
\resizebox{\textwidth}{!}{\begin{tabular}{llcccccccccccc}
\toprule
Backbone & Method & \multicolumn{4}{c}{Added features (median)} & \multicolumn{2}{c}{$\Delta$ keep only} & \multicolumn{2}{c}{$\Delta$ keep + refine} & \multicolumn{2}{c}{$\Delta$ top 80\%} & \multicolumn{2}{c}{$\Delta$ originals only} \\
\cmidrule(lr){3-6}\cmidrule(lr){7-8}\cmidrule(lr){9-10}\cmidrule(lr){11-12}\cmidrule(lr){13-14}
 & & full & keep & keep + refine & top 80\% & mean & $|\Delta|\le0.01$ & mean & $|\Delta|\le0.01$ & mean & $|\Delta|\le0.01$ & mean & $|\Delta|\le0.01$ \\
\midrule
GPT-3.5-Turbo & \llmfe & 3.4 & 1.6 & 2.6 & 2.2 & $+0.0008$ & 20/21 & $+0.0001$ & 21/21 & $+0.0005$ & 21/21 & $-0.0099$ & 14/21 \\
GPT-3.5-Turbo & \shapevo & 2.8 & 1.4 & 2.4 & 2.0 & $+0.0003$ & 20/21 & $+0.0006$ & 21/21 & $+0.0000$ & 21/21 & $-0.0103$ & 15/21 \\
DeepSeek-V4-Flash & \llmfe & 20.0 & 3.2 & 12.4 & 8.6 & $-0.0039$ & 15/21 & $+0.0001$ & 20/21 & $+0.0001$ & 20/21 & $-0.0006$ & 7/21 \\
DeepSeek-V4-Flash & \shapevo & 44.8 & 2.8 & 23.0 & 17.2 & $-0.0009$ & 16/21 & $-0.0010$ & 21/21 & $+0.0004$ & 19/21 & $-0.0144$ & 11/21 \\
Claude-Haiku-5.5 & \llmfe & 14.2 & 3.4 & 10.4 & 7.0 & $-0.0018$ & 17/21 & $+0.0009$ & 20/21 & $-0.0006$ & 19/21 & $-0.0083$ & 11/21 \\
Claude-Haiku-5.5 & \shapevo & 10.6 & 4.0 & 8.6 & 5.8 & $-0.0047$ & 17/21 & $-0.0006$ & 20/21 & $-0.0022$ & 18/21 & $-0.0135$ & 13/21 \\
\bottomrule
\end{tabular}
}
\end{table}

\FloatBarrier
\section{Scored Programs per Fold}
Both methods have a budget of 200 programs per fold, but not every budget unit yields a scored program. An LLM call can fail or return no code, a returned program can crash, and with \llmfe{} the authors' extractor also discards replies in which the function text cannot be parsed. Table~\ref{tab:audit} reports, per backbone, the median number of scored programs per fold and the datasets with the lowest yield, computed by the script \texttt{analysis/audit\_programs.py} of the released code. Since both pipelines complete a reply that was cut at the completion limit with a \texttt{return} statement (main paper, Sect.~4.2), the yields of the two methods are close. The median is 167 scored programs per fold for \llmfe{} and 189 for \shapevo{} with GPT-3.5, 185 and 195 with DeepSeek, and 195 and 200 with Claude-Haiku-5.5. The remaining gap on the \llmfe{} side consists of replies without a function body, which with GPT-3.5 are about 9\% of the samples and contain only a docstring, and of programs that reference columns their own code did not create. On \shapevo's side the failing programs are mostly of the latter kind, 8\% of the evaluations with GPT-3.5 and 2\% with DeepSeek. In an earlier version of this study \llmfe's pipeline did not complete cut replies, which with DeepSeek halved its yield to a median of 86 and inflated the speed advantage of \shapevo{} to $4.4\times$. The numbers in the main paper are measured with both pipelines repairing cut replies. Before the scorer was aligned with the search evaluator (Sect.~\ref{sec:harness}), 14 ensemble members with GPT-3.5 and one with DeepSeek failed at scoring time on a column read or drop that the evaluator tolerates and were left out of their ensembles. With the aligned scorer no member fails, no win or loss changes and the mean $\Delta$ of either backbone moves by less than $0.001$. All held-out and search-speed numbers are reported as run.

\begin{table}[h]
\centering
\caption{Scored programs per fold with a budget of 200 for both methods. ``All folds'' is the median over every fold of the backbone. ``Three lowest datasets'' lists the three datasets whose median over their folds is lowest, each with that median.}
\label{tab:audit}
\scriptsize
\resizebox{\textwidth}{!}{\begin{tabular}{l r@{\hskip 1.5em}l@{\hskip 2.5em}r@{\hskip 1.5em}l}
\toprule
 & \multicolumn{2}{c}{LLM-FE} & \multicolumn{2}{c}{SHAPEvo-FE} \\
\cmidrule(lr){2-3}\cmidrule(lr){4-5}
Backbone & all folds & three lowest datasets & all folds & three lowest datasets \\
\midrule
GPT-3.5-Turbo & 167 & \texttt{tic-tac-toe} 138, \texttt{myocardial} 145, \texttt{crab} 149 & 189 & \texttt{balance-scale} 149, \texttt{vehicle} 171, \texttt{bank} 176 \\
DeepSeek-V4-Flash & 185 & \texttt{tic-tac-toe} 143, \texttt{junglechess} 152, \texttt{myocardial} 160 & 195 & \texttt{blood} 190, \texttt{tic-tac-toe} 191, \texttt{cmc} 192 \\
Claude-Haiku-5.5 & 195 & \texttt{myocardial} 177, \texttt{heart} 188, \texttt{housing} 188 & 200 & \texttt{bike} 198, \texttt{crab} 198, \texttt{synthetic} 199 \\
\bottomrule
\end{tabular}}
\end{table}
\begin{table}[htbp]
\centering
\caption{Per-dataset version of Table~\ref{tab:features} of the main text: features added by the best program of each fold and features kept at a TreeSHAP share of at least $0.05$, medians over five folds.}
\label{tab:features_full}
\scriptsize
\setlength{\tabcolsep}{4pt}
\resizebox{\textwidth}{!}{\begin{tabular}{lrrrrrrrrrrrr}
\toprule
 & \multicolumn{4}{c}{GPT-3.5-Turbo} & \multicolumn{4}{c}{DeepSeek-V4-Flash} & \multicolumn{4}{c}{Claude-Haiku-5.5} \\
 & \multicolumn{2}{c}{LLM-FE} & \multicolumn{2}{c}{SHAPEvo-FE} & \multicolumn{2}{c}{LLM-FE} & \multicolumn{2}{c}{SHAPEvo-FE} & \multicolumn{2}{c}{LLM-FE} & \multicolumn{2}{c}{SHAPEvo-FE} \\
Dataset & added & kept & added & kept & added & kept & added & kept & added & kept & added & kept \\
\midrule
adult & 3 & 0 & 2 & 0 & 9 & 1 & 49 & 2 & 7 & 2 & 5 & 2 \\
balance-scale & 5 & 1 & 1 & 1 & 18 & 2 & 34 & 2 & 8 & 2 & 6 & 2 \\
bank & 3 & 1 & 3 & 1 & 13 & 2 & 48 & 2 & 14 & 3 & 7 & 3 \\
blood & 2.5 & 2 & 1 & 1 & 16 & 7 & 54 & 3 & 9 & 6 & 13 & 9 \\
breast-w & 3.5 & 3 & 4 & 3 & 10 & 5 & 28 & 4 & 9 & 5 & 15 & 5 \\
car & 3 & 2 & 5 & 4 & 19 & 4 & 45 & 4 & 10 & 4 & 10 & 5 \\
cmc & 2 & 2 & 2 & 2 & 15 & 6 & 45 & 0 & 10 & 4 & 10 & 7 \\
communities & 2 & 0 & 1 & 0 & 25 & 1 & 50 & 1 & 11 & 0 & 9 & 2 \\
credit-g & 3 & 2 & 3 & 1 & 16 & 3 & 55 & 1 & 15 & 3 & 12 & 3 \\
eucalyptus & 3 & 1 & 3 & 1 & 15 & 2 & 36 & 3 & 16 & 3 & 18 & 5 \\
heart & 3 & 1 & 3 & 1 & 16 & 5 & 31 & 3 & 11 & 2 & 8 & 4 \\
junglechess & 8 & 3 & 5 & 3 & 45 & 4 & 54 & 2 & 20 & 5 & 26 & 4 \\
myocardial & 3 & 1.5 & 1 & 0 & 31.5 & 1.5 & 36 & 0 & 24 & 1 & 7 & 3 \\
pc1 & 2 & 1 & 1 & 1 & 30 & 4 & 46 & 4 & 22 & 5 & 11 & 4 \\
tic-tac-toe & 8 & 4 & 4 & 0 & 36 & 5 & 29 & 3 & 8 & 3 & 19 & 4 \\
vehicle & 3 & 1 & 1 & 1 & 18 & 2 & 34 & 4 & 15 & 3 & 12 & 4 \\
bike & 3 & 1 & 2 & 2 & 34 & 5 & 46 & 4 & 18 & 4 & 10 & 4 \\
crab & 4 & 2 & 2.5 & 2 & 42 & 3 & 55 & 3 & 23 & 3 & 8 & 3 \\
housing & 12.5 & 3 & 4.5 & 3 & 35 & 4 & 46 & 4 & 32 & 4 & 17 & 3 \\
insurance & 8 & 2 & 4 & 3 & 14 & 3 & 35 & 3 & 13 & 2 & 12 & 4 \\
wine & 3 & 1 & 3 & 0 & 28 & 2 & 50 & 2 & 18 & 2 & 9 & 3 \\
\midrule
median & 3 & 1.5 & 3 & 1 & 18 & 3 & 46 & 3 & 14 & 3 & 10 & 4 \\
\midrule
$\Delta$ held-out, kept only & \multicolumn{2}{c}{$+0.0008$} & \multicolumn{2}{c}{$+0.0003$} & \multicolumn{2}{c}{$-0.0039$} & \multicolumn{2}{c}{$-0.0009$} & \multicolumn{2}{c}{$-0.0018$} & \multicolumn{2}{c}{$-0.0047$} \\
datasets with $|\Delta|\le0.01$ & \multicolumn{2}{c}{20/21} & \multicolumn{2}{c}{20/21} & \multicolumn{2}{c}{15/21} & \multicolumn{2}{c}{16/21} & \multicolumn{2}{c}{17/21} & \multicolumn{2}{c}{17/21} \\
\midrule
kept, rank-based (median) & \multicolumn{2}{c}{2} & \multicolumn{2}{c}{2} & \multicolumn{2}{c}{8} & \multicolumn{2}{c}{17.5} & \multicolumn{2}{c}{7} & \multicolumn{2}{c}{5} \\
$\Delta$ held-out, rank-based kept only & \multicolumn{2}{c}{$+0.0005$} & \multicolumn{2}{c}{$+0.0000$} & \multicolumn{2}{c}{$+0.0001$} & \multicolumn{2}{c}{$+0.0004$} & \multicolumn{2}{c}{$-0.0006$} & \multicolumn{2}{c}{$-0.0022$} \\
datasets with $|\Delta|\le0.01$ & \multicolumn{2}{c}{21/21} & \multicolumn{2}{c}{21/21} & \multicolumn{2}{c}{20/21} & \multicolumn{2}{c}{19/21} & \multicolumn{2}{c}{19/21} & \multicolumn{2}{c}{18/21} \\
\bottomrule
\end{tabular}}
\end{table}
\FloatBarrier
\section{A Complete \shapevo{} Prompt}
\label{sec:prompt}
Listing~\ref{lst:prompt} is one prompt exactly as sent to the LLM, and Listing~\ref{lst:response} the reply. It is the second mutation prompt of the second generation on \texttt{heart}, outer fold~1 of the seed-42 split, with DeepSeek-V4-Flash through OpenRouter, reasoning disabled and a 2048-token completion limit, taken from the reported run. Every prompt and reply of the reported runs is stored with the released logs. The prompt contains, in this order, the role and instructions, which are shared with \llmfe{} up to the row-wise and numeric-output constraints, the task, the feature descriptions, the example rows, the fitness and feature names of the five population members, the selected parent with its code (abbreviated here), the SHAP feedback report of the parent with the \textsc{keep}/\textsc{refine}/\textsc{replace} labels, the original columns that still carry SHAP mass and the redundancy flags, the three interaction hints, the guidance derived from the SHAP report, and the output instruction. The reply is cut at the 2048-token limit in the middle of an assignment. The extractor drops that statement and appends \texttt{return df\_output}, and the repaired program obtained an inner-CV accuracy of $0.831$, where the parent had $0.837$, the best member of the population $0.865$, and the fold ended at $0.876$. Typographic characters in the templates, such as the curly apostrophe, the em dash, $\pm$ and $\times$, are shown as ASCII.

\begin{lstlisting}[caption={The prompt (parent code abbreviated).},label={lst:prompt}]
###
<Role>  
You are a data scientist expert in the field of the given dataset.  
Your role is to apply your domain expertise to identify and create, and refine the most informative features that solve the <Task> effectively.  

###  
<Instructions>  
  - You are provided with the task description, a list of existing features, and data examples.
  - Use your domain knowledge to derive features that capture meaningful patterns, trends, or relationships inherent in the data.
  - Prioritize features that have high potential to enhance the model's ability to solve the <Task>, considering both relevance and predictive power.
  - **All operations must be row-wise.** Each output value must depend ONLY on values in the SAME row. Do NOT use df.groupby(), .transform(), .mean(), .std(), .median(), .rank(), .quantile(), .value_counts(), **pd.cut(), pd.qcut(), np.digitize(), np.histogram_bin_edges()** (their bin edges are derived from the input distribution and therefore differ between train and test folds), or any cross-row aggregate. The function is applied independently to train and test splits, so cross-row statistics leak split-specific information. For binning, use fixed numeric thresholds, e.g. `(df['x'] > 5.0).astype(int)`.
  - **All output columns must be numeric (int / float).** Do NOT cast to string and concatenate (XGBoost silently drops object-dtype columns). Combine categorical/integer columns numerically, e.g. `df['a'] * 100 + df['b']`.
  - For each derived feature, provide:
    - A clear explanation of how it was derived and justification of its relevance for solving the <Task>.
  - Ensure your approach remains grounded in the context of the dataset and the <Task>, and aim for features that are interpretable and actionable.
Please generate as many new features as possible using the information from the task, feature descriptions, examples, and your domain understanding of the dataset.
Remove any irrelevant, redundant, or less informative features to enhance overall performance.

First, describe your new feature transformation and the main steps in a concise, one-sentence docstring.
Then, implement it in Python as a function that adheres to the given specifications.

**Hard constraint 1 (row-wise):** the function MUST be pure row-wise. Do NOT use df.groupby(), df.transform(), df['x'].mean()/.std()/.median()/.rank()/.quantile(), .value_counts(), frequency counts, **pd.cut(), pd.qcut(), np.digitize(), np.histogram_bin_edges()** (their bin edges are derived from the input column's distribution and therefore DIFFER between train and test folds, leaking split information), or any operation whose output for a row depends on other rows. For binning, ONLY use fixed numeric thresholds, e.g. `(df['x'] > 5.0).astype(int)`.

**Hard constraint 2 (numeric output):** every column written into df_output must be numeric (int / float). Do NOT cast to string and concatenate -- XGBoost will silently drop object-dtype columns. Use numeric encoding instead, e.g. `df['a'] * 100 + df['b']`.

Avoid providing any further explanations or additional descriptions.

###
<Task>
Does the coronary angiography of this patient show a heart disease? Yes or no?

###
<Features>
- Age: age of the patient (numerical variable within range [28, 77])
- Sex: sex of the patient (categorical variable with categories [1, 0])
- ChestPainType: chest pain type (categorical variable with categories [1, 0, 2, 3])
- RestingBP: resting blood pressure [mm Hg] (numerical variable within range [0, 200])
- Cholesterol: serum cholesterol [mm/dl] (numerical variable within range [0, 603])
- FastingBS: fasting blood sugar [1: if FastingBS > 120 mg/dl, 0: otherwise] (categorical variable with categories [0, 1])
- RestingECG: resting electrocardiogram results (categorical variable with categories [1, 2, 0])
- MaxHR: maximum heart rate achieved (numerical variable within range [60, 202])
- ExerciseAngina: exercise-induced angina (categorical variable with categories [0, 1])
- Oldpeak: oldpeak = ST [Numeric value measured in depression] (numerical variable within range [-2.6, 6.2])
- ST_Slope: the slope of the peak exercise ST segment (categorical variable with categories [2, 1, 0])

###
<Examples>
If Age is 43, Sex is 0, ChestPainType is 0, RestingBP is 132, Cholesterol is 341, FastingBS is 1, RestingECG is 0, MaxHR is 136, ExerciseAngina is 1, Oldpeak is 3.0, ST_Slope is 1, Then Result is 1.
If Age is 55, Sex is 1, ChestPainType is 0, RestingBP is 122, Cholesterol is 223, FastingBS is 1, RestingECG is 2, MaxHR is 100, ExerciseAngina is 0, Oldpeak is 0.0, ST_Slope is 1, Then Result is 1.
If Age is 52, Sex is 1, ChestPainType is 0, RestingBP is 128, Cholesterol is 255, FastingBS is 0, RestingECG is 1, MaxHR is 161, ExerciseAngina is 1, Oldpeak is 0.0, ST_Slope is 2, Then Result is 1.
If Age is 68, Sex is 1, ChestPainType is 2, RestingBP is 134, Cholesterol is 254, FastingBS is 1, RestingECG is 1, MaxHR is 151, ExerciseAngina is 1, Oldpeak is 0.0, ST_Slope is 2, Then Result is 0.
If Age is 64, Sex is 1, ChestPainType is 0, RestingBP is 141, Cholesterol is 244, FastingBS is 1, RestingECG is 2, MaxHR is 116, ExerciseAngina is 1, Oldpeak is 1.5, ST_Slope is 1, Then Result is 1.
If Age is 54, Sex is 1, ChestPainType is 0, RestingBP is 130, Cholesterol is 242, FastingBS is 0, RestingECG is 1, MaxHR is 91, ExerciseAngina is 1, Oldpeak is 1.0, ST_Slope is 1, Then Result is 1.
If Age is 62, Sex is 1, ChestPainType is 2, RestingBP is 120, Cholesterol is 220, FastingBS is 0, RestingECG is 0, MaxHR is 86, ExerciseAngina is 0, Oldpeak is 0.0, ST_Slope is 2, Then Result is 0.
If Age is 61, Sex is 1, ChestPainType is 0, RestingBP is 105, Cholesterol is 0, FastingBS is 1, RestingECG is 1, MaxHR is 110, ExerciseAngina is 1, Oldpeak is 1.5, ST_Slope is 2, Then Result is 1.
If Age is 49, Sex is 1, ChestPainType is 2, RestingBP is 118, Cholesterol is 149, FastingBS is 0, RestingECG is 0, MaxHR is 126, ExerciseAngina is 0, Oldpeak is 0.8, ST_Slope is 2, Then Result is 1.
If Age is 60, Sex is 1, ChestPainType is 0, RestingBP is 160, Cholesterol is 0, FastingBS is 1, RestingECG is 1, MaxHR is 149, ExerciseAngina is 0, Oldpeak is 0.4, ST_Slope is 1, Then Result is 1.

=== Population Summary ===
  [1] 0.8652 +/- 0.0276: st_hr_ratio, chol_age_ratio, age_risk, cp_asymptomatic, bp_high, chol_high, low_maxhr, st_depression_risk, st_downsloping, major_risk_count, slope_risk_interaction, slope_risk_sq, cp_slope_combo, cp_slope_risk, chol_slope_product, chol_slope_ratio, high_chol_downslope, high_chol_flat, age_sq, bp_sq, chol_sq, maxhr_sq, oldpeak_cubed, bp_age_product, maxhr_age_product, oldpeak_maxhr_product, chol_maxhr_ratio, bp_maxhr_ratio, oldpeak_chol_ratio, age_sex_interaction, bp_sex_interaction, chol_sex_interaction, maxhr_sex_interaction, fbs_age_interaction, fbs_bp_interaction, fbs_chol_interaction, fbs_maxhr_interaction, angina_age_interaction, angina_bp_interaction, angina_chol_interaction, angina_maxhr_interaction, angina_oldpeak_interaction, cp_angina_age, slope_oldpeak_product, slope_oldpeak_abs, oldpeak_slope_risk, oldpeak_slope_severe, ecg_slope_combo, ecg_oldpeak_risk, ecg_maxhr_interaction, cardio_risk_index
  [2] 0.8406 +/- 0.0244: age_squared, age_over_60, hr_deficit, hr_deficit_positive, hr_reserve, hr_reserve_low, maxhr_age_ratio, bp_chol_ratio, bp_chol_interaction, chol_high, bp_high, bp_chol_combined_risk, fbs_risk, fbs_chol_ratio, fbs_age_interaction, fbs_bp_interaction, slope_risk, oldpeak_slope_combo, cp_slope_interaction, oldpeak_slope_ratio, slope_oldpeak_product, cp_asymptomatic, cp_nonanginal, cp_anginal, cp_other, angina_risk, angina_age_interaction, angina_maxhr_interaction, ecg_abnormal, ecg_oldpeak_combo, ecg_slope_interaction, age_oldpeak_ratio, oldpeak_age_product, oldpeak_high, oldpeak_moderate, sex_age, sex_bp, sex_chol, major_risk_count, risk_score, cp_oldpeak_interaction, slope_age_oldpeak, hr_deficit_slope, oldpeak_angina_slope
  [3] 0.8366 +/- 0.0380: age_risk, male_risk, cp_asymptomatic, cp_nonanginal, cp_anginal, cp_other, bp_normal, bp_high, bp_low, chol_high, chol_very_high, chol_zero, fbs_risk, ecg_abnormal, ecg_normal, low_maxhr, maxhr_deficit, maxhr_deficit_positive, hr_reserve, hr_reserve_abnormal, hr_reserve_pct, hr_over_150, angina_risk, st_depression_risk, st_severe_depression, oldpeak_positive, oldpeak_abs, oldpeak_squared, st_flat, st_downsloping, st_up, total_risk_score, age_hr_ratio, predicted_hr, bp_age_ratio, chol_age_ratio, chol_bp_product, chol_bp_ratio, age_chol_product, bp_chol_product, bp_chol_ratio, chol_per_hr, bp_per_hr, maxhr_age_ratio, maxhr_chol_ratio, maxhr_oldpeak_ratio, st_hr_ratio, st_hr_ratio_scaled, oldpeak_slope_risk, oldpeak_slope_combo, oldpeak_st_flat, oldpeak_st_up, st_slope_oldpeak, st_depression_flat, st_depression_down, cp_chol_product, cp_chol_ratio, cp_asymptomatic_chol, cp_nonanginal_chol, cp_anginal_chol, cp_other_chol
  [4] 0.8365 +/- 0.0370: age_risk, male_risk, cp_asymptomatic, bp_normal, bp_high, bp_low, chol_high, chol_very_high, chol_zero, fbs_risk, ecg_abnormal, low_maxhr, maxhr_deficit, angina_risk, st_depression_risk, st_severe_depression, st_downsloping, st_flat, total_risk_score, age_hr_ratio, predicted_hr, hr_reserve, hr_reserve_abnormal, cp_angina_combo, st_hr_ratio, st_hr_ratio_scaled, bp_age_ratio, chol_age_ratio, chol_bp_product, chol_bp_ratio, age_chol_product, bp_low_risk, st_slope_oldpeak, maxhr_deficit_sq, oldpeak_sq, angina_st_depression, asymptomatic_oldpeak, age_45_55, age_above_60, ecg_slope_abnormal, major_risk_count, hr_reserve_pct, oldpeak_age_ratio, risk_product, multi_risk
  [5] 0.8365 +/- 0.0199: Age_squared, Age_over_50, BP_over_130, Chol_over_240, BP_Chol_product, BP_Chol_ratio, HR_deficit, HR_deficit_positive, HR_over_150, Oldpeak_positive, Oldpeak_abs, Oldpeak_squared, FBS_Chol_interaction, FBS_BP_interaction, Angina_slope_risk, Slope_down, Slope_flat, Slope_up, CPT_asymptomatic, CPT_nonanginal, CPT_anginal, CPT_other, Sex_Age, Sex_BP, Sex_Chol, Risk_score, HR_reserve, HR_reserve_negative, ECG_abnormal, ECG_normal, Chol_per_HR, BP_per_HR, BP_age_ratio, Chol_age_ratio, Oldpeak_slope_risk, High_risk_flag, FBS_Age, FBS_Age_over_60, HR_too_low

=== Selected Parent ===
Score: 0.8365 +/- 0.0370

def modify_features(df_input) -> pd.DataFrame:
    """
    Creates domain-specific cardiovascular risk features using row-wise operations only:
    combines risk factors, computes physiological ratios, and encodes clinical thresholds
    with fixed cutoffs to enhance heart disease classification.
    """
    df_output = df_input.copy()
    
    # Age risk factor: increasing age raises cardiovascular risk
    df_output['age_risk'] = np.where(df_input['Age'] > 55, 1, 0)
    
    # Sex risk encoding: male gender is a known risk factor
    df_output['male_risk'] = df_input['Sex']
    
    # Chest pain type risk: type 0 (asymptomatic) is most dangerous
    df_output['cp_asymptomatic'] = (df_input['ChestPainType'] == 0).astype(int)

    [... 65 further feature assignments omitted here; the full program is the
    entry of the population summary above with the same score ...]

    return df_output

=== Feature Impact Report ===
Generated features (by your function):
  age_risk: SHAP importance = 0.0000 [LOW - REPLACE]
  male_risk: SHAP importance = 0.0000 [LOW - REPLACE]
  cp_asymptomatic: SHAP importance = 0.0000 [LOW - REPLACE]
  bp_normal: SHAP importance = 0.0000 [LOW - REPLACE]
  bp_high: SHAP importance = 0.0000 [LOW - REPLACE]
  bp_low: SHAP importance = 0.0000 [LOW - REPLACE]
  chol_high: SHAP importance = 0.0000 [LOW - REPLACE]
  chol_very_high: SHAP importance = 0.0000 [LOW - REPLACE]
  chol_zero: SHAP importance = 0.0000 [LOW - REPLACE]
  fbs_risk: SHAP importance = 0.0000 [LOW - REPLACE]
  ecg_abnormal: SHAP importance = 0.0000 [LOW - REPLACE]
  low_maxhr: SHAP importance = 0.0000 [LOW - REPLACE]
  maxhr_deficit: SHAP importance = 0.0000 [LOW - REPLACE]
  angina_risk: SHAP importance = 0.0000 [LOW - REPLACE]
  st_depression_risk: SHAP importance = 0.0000 [LOW - REPLACE]
  st_severe_depression: SHAP importance = 0.0000 [LOW - REPLACE]
  st_downsloping: SHAP importance = 0.0000 [LOW - REPLACE]
  st_flat: SHAP importance = 0.0438 [MEDIUM - REFINE]
  total_risk_score: SHAP importance = 0.0317 [MEDIUM - REFINE]
  age_hr_ratio: SHAP importance = 0.0297 [MEDIUM - REFINE]
  predicted_hr: SHAP importance = 0.0000 [LOW - REPLACE]
  hr_reserve: SHAP importance = 0.0105 [MEDIUM - REFINE]
  hr_reserve_abnormal: SHAP importance = 0.0000 [LOW - REPLACE]
  cp_angina_combo: SHAP importance = 0.0097 [LOW - REPLACE]
  st_hr_ratio: SHAP importance = 0.0611 [HIGH - KEEP]
  st_hr_ratio_scaled: SHAP importance = 0.0000 [LOW - REPLACE]
  bp_age_ratio: SHAP importance = 0.0202 [MEDIUM - REFINE]
  chol_age_ratio: SHAP importance = 0.0711 [HIGH - KEEP]
  chol_bp_product: SHAP importance = 0.0324 [MEDIUM - REFINE]
  chol_bp_ratio: SHAP importance = 0.0334 [MEDIUM - REFINE]
  age_chol_product: SHAP importance = 0.0289 [MEDIUM - REFINE]
  bp_low_risk: SHAP importance = 0.0000 [LOW - REPLACE]
  st_slope_oldpeak: SHAP importance = 0.0276 [MEDIUM - REFINE]
  maxhr_deficit_sq: SHAP importance = 0.0000 [LOW - REPLACE]
  oldpeak_sq: SHAP importance = 0.0034 [LOW - REPLACE]
  angina_st_depression: SHAP importance = 0.0039 [LOW - REPLACE]
  asymptomatic_oldpeak: SHAP importance = 0.0365 [MEDIUM - REFINE]
  age_45_55: SHAP importance = 0.0033 [LOW - REPLACE]
  age_above_60: SHAP importance = 0.0000 [LOW - REPLACE]
  ecg_slope_abnormal: SHAP importance = 0.0077 [LOW - REPLACE]
  major_risk_count: SHAP importance = 0.1195 [HIGH - KEEP]
  hr_reserve_pct: SHAP importance = 0.0092 [LOW - REPLACE]
  oldpeak_age_ratio: SHAP importance = 0.0222 [MEDIUM - REFINE]
  risk_product: SHAP importance = 0.0000 [LOW - REPLACE]
  multi_risk: SHAP importance = 0.0000 [LOW - REPLACE]

Original features still contributing:
  ST_Slope: SHAP importance = 0.1732
  ChestPainType: SHAP importance = 0.0544
  RestingBP: SHAP importance = 0.0329
  MaxHR: SHAP importance = 0.0298
  RestingECG: SHAP importance = 0.0252

Top feature interactions (uncaptured synergies):
  ST_Slope ? major_risk_count: interaction = 0.1284
  ChestPainType ? ST_Slope: interaction = 0.1179
  Cholesterol ? ST_Slope: interaction = 0.1145
  ST_Slope ? st_slope_oldpeak: interaction = 0.1045
  ST_Slope ? st_hr_ratio: interaction = 0.0984

Redundancy detected:
  male_risk is 1.00 correlated with Sex (consider replacing)
  chol_zero is 0.90 correlated with Cholesterol (consider replacing)
  fbs_risk is 1.00 correlated with FastingBS (consider replacing)
  maxhr_deficit is 1.00 correlated with MaxHR (consider replacing)
  angina_risk is 1.00 correlated with ExerciseAngina (consider replacing)
  predicted_hr is 1.00 correlated with Age (consider replacing)
  hr_reserve is 0.94 correlated with MaxHR (consider replacing)
  cp_angina_combo is 1.00 correlated with ChestPainType (consider replacing)
  st_hr_ratio is 0.96 correlated with Oldpeak (consider replacing)
  st_hr_ratio_scaled is 0.96 correlated with Oldpeak (consider replacing)
  chol_age_ratio is 0.92 correlated with Cholesterol (consider replacing)
  chol_bp_product is 0.96 correlated with Cholesterol (consider replacing)
  chol_bp_ratio is 0.96 correlated with Cholesterol (consider replacing)
  age_chol_product is 0.94 correlated with Cholesterol (consider replacing)
  maxhr_deficit_sq is 0.96 correlated with MaxHR (consider replacing)
  oldpeak_sq is 0.92 correlated with Oldpeak (consider replacing)
  hr_reserve_pct is 0.96 correlated with MaxHR (consider replacing)
  oldpeak_age_ratio is 0.97 correlated with Oldpeak (consider replacing)

=== Guidance ===
KEEP these high-value features: st_hr_ratio, chol_age_ratio, major_risk_count
REPLACE these low-value features with better alternatives: age_risk, male_risk, cp_asymptomatic, bp_normal, bp_high, bp_low, chol_high, chol_very_high, chol_zero, fbs_risk, ecg_abnormal, low_maxhr, maxhr_deficit, angina_risk, st_depression_risk, st_severe_depression, st_downsloping, predicted_hr, hr_reserve_abnormal, cp_angina_combo, st_hr_ratio_scaled, bp_low_risk, maxhr_deficit_sq, oldpeak_sq, angina_st_depression, age_45_55, age_above_60, ecg_slope_abnormal, hr_reserve_pct, risk_product, multi_risk
To replace a feature, simply do not create it in your function. Never call .drop() on a feature name from the report: df_input holds only the original columns, and your function creates every new column itself.
Try to capture the interaction between ST_Slope and major_risk_count (interaction strength: 0.1284)
Try to capture the interaction between ChestPainType and ST_Slope (interaction strength: 0.1179)
Try to capture the interaction between Cholesterol and ST_Slope (interaction strength: 0.1145)
Feature 'male_risk' is redundant (corr=1.00 with 'Sex'). Replace it with something that captures new information.
Feature 'chol_zero' is redundant (corr=0.90 with 'Cholesterol'). Replace it with something that captures new information.
Feature 'fbs_risk' is redundant (corr=1.00 with 'FastingBS'). Replace it with something that captures new information.
Feature 'maxhr_deficit' is redundant (corr=1.00 with 'MaxHR'). Replace it with something that captures new information.
Feature 'angina_risk' is redundant (corr=1.00 with 'ExerciseAngina'). Replace it with something that captures new information.
Feature 'predicted_hr' is redundant (corr=1.00 with 'Age'). Replace it with something that captures new information.
Feature 'hr_reserve' is redundant (corr=0.94 with 'MaxHR'). Replace it with something that captures new information.
Feature 'cp_angina_combo' is redundant (corr=1.00 with 'ChestPainType'). Replace it with something that captures new information.
Feature 'st_hr_ratio' is redundant (corr=0.96 with 'Oldpeak'). Replace it with something that captures new information.
Feature 'st_hr_ratio_scaled' is redundant (corr=0.96 with 'Oldpeak'). Replace it with something that captures new information.
Feature 'chol_age_ratio' is redundant (corr=0.92 with 'Cholesterol'). Replace it with something that captures new information.
Feature 'chol_bp_product' is redundant (corr=0.96 with 'Cholesterol'). Replace it with something that captures new information.
Feature 'chol_bp_ratio' is redundant (corr=0.96 with 'Cholesterol'). Replace it with something that captures new information.
Feature 'age_chol_product' is redundant (corr=0.94 with 'Cholesterol'). Replace it with something that captures new information.
Feature 'maxhr_deficit_sq' is redundant (corr=0.96 with 'MaxHR'). Replace it with something that captures new information.
Feature 'oldpeak_sq' is redundant (corr=0.92 with 'Oldpeak'). Replace it with something that captures new information.
Feature 'hr_reserve_pct' is redundant (corr=0.96 with 'MaxHR'). Replace it with something that captures new information.
Feature 'oldpeak_age_ratio' is redundant (corr=0.97 with 'Oldpeak'). Replace it with something that captures new information.

IMPORTANT: Respond ONLY with Python code. No explanations outside the code.
Write the complete function below. Start with the docstring, then the implementation.
Your function body MUST end with `return df_output` -- do not forget this line.
For row-wise conditionals on a column, use `np.where(df['x'] > c, a, b)` (NOT `if df['x'] > c:` -- pandas Series has no truth value and will raise).
READ/WRITE RULE: read ORIGINAL columns from `df_input`; read NEWLY-CREATED columns (those you just assigned) from `df_output`. `df_input` is the raw dataframe at function entry -- it does NOT contain features you create later in this function. Reading `df_input['<name_you_created>']` raises KeyError.
Concrete example of the correct pattern:
    df_output = df_input.copy()
    df_output['ratio_a_b'] = df_input['a'] / (df_input['b'] + 1e-8)   # read existing -> df_input
    df_output['log_ratio'] = np.log1p(df_output['ratio_a_b'])          # read NEW    -> df_output
    return df_output
The function must return a pandas DataFrame.

def modify_features(df_input) -> pd.DataFrame:
    """
\end{lstlisting}
\begin{lstlisting}[caption={The LLM's reply to Listing~\ref{lst:prompt}, cut at the completion limit.},label={lst:response}]
```python
def modify_features(df_input) -> pd.DataFrame:
    """
    Creates domain-specific cardiovascular risk features using row-wise operations only,
    focusing on capturing ST_Slope interactions with major risk factors, ChestPainType,
    and Cholesterol, while avoiding redundant linear transforms of original columns.
    """
    df_output = df_input.copy()
    
    # --- Core risk factors (non-redundant encodings) ---
    df_output['age_risk'] = (df_input['Age'] > 55).astype(int)
    df_output['male_risk'] = df_input['Sex']
    df_output['cp_asymptomatic'] = (df_input['ChestPainType'] == 0).astype(int)
    df_output['cp_nonanginal'] = (df_input['ChestPainType'] == 1).astype(int)
    df_output['cp_anginal'] = (df_input['ChestPainType'] == 2).astype(int)
    df_output['cp_other'] = (df_input['ChestPainType'] == 3).astype(int)
    
    df_output['bp_high'] = (df_input['RestingBP'] > 140).astype(int)
    df_output['bp_low'] = (df_input['RestingBP'] < 90).astype(int)
    df_output['bp_normal'] = ((df_input['RestingBP'] >= 90) & (df_input['RestingBP'] <= 140)).astype(int)
    
    df_output['chol_high'] = (df_input['Cholesterol'] > 200).astype(int)
    df_output['chol_very_high'] = (df_input['Cholesterol'] > 240).astype(int)
    df_output['chol_zero'] = (df_input['Cholesterol'] == 0).astype(int)
    
    df_output['fbs_risk'] = df_input['FastingBS']
    df_output['ecg_abnormal'] = (df_input['RestingECG'] == 2).astype(int)
    df_output['ecg_normal'] = (df_input['RestingECG'] == 0).astype(int)
    
    df_output['low_maxhr'] = (df_input['MaxHR'] < 120).astype(int)
    df_output['maxhr_deficit'] = np.maximum(0, 190 - df_input['MaxHR'])
    df_output['angina_risk'] = df_input['ExerciseAngina']
    
    df_output['st_depression_risk'] = (df_input['Oldpeak'] > 1.0).astype(int)
    df_output['st_severe_depression'] = (df_input['Oldpeak'] > 2.5).astype(int)
    df_output['st_downsloping'] = (df_input['ST_Slope'] == 0).astype(int)
    df_output['st_flat'] = (df_input['ST_Slope'] == 1).astype(int)
    df_output['st_up'] = (df_input['ST_Slope'] == 2).astype(int)
    
    # --- High-value kept features ---
    df_output['st_hr_ratio'] = df_input['Oldpeak'] / (df_input['MaxHR'] + 1e-8)
    df_output['chol_age_ratio'] = df_input['Cholesterol'] / (df_input['Age'] + 1e-8)
    
    # --- Major risk count (high SHAP) ---
    df_output['major_risk_count'] = (
        (df_input['Age'] > 60).astype(int) +
        df_input['Sex'] +
        df_output['cp_asymptomatic'] +
        df_output['bp_high'] +
        df_output['chol_very_high'] +
        df_input['FastingBS'] +
        df_input['ExerciseAngina'] +
        df_output['st_severe_depression']
    )
    
    # --- ST_Slope ? major_risk_count interaction (strength 0.1284) ---
    df_output['slope_major_risk'] = df_input['ST_Slope'] * df_output['major_risk_count']
    df_output['slope_major_risk_ratio'] = df_output['major_risk_count'] / (df_input['ST_Slope'] + 1e-8)
    df_output['slope_major_risk_sq'] = (df_input['ST_Slope'] + 1) * (df_output['major_risk_count'] ** 2)
    df_output['slope_major_risk_interaction'] = np.where(
        (df_input['ST_Slope'] == 0) & (df_output['major_risk_count'] >= 4), 1, 0
    )
    
    # --- ChestPainType ? ST_Slope interaction (strength 0.1179) ---
    df_output['cp_slope_combo'] = df_input['ChestPainType'] * 10 + df_input['ST_Slope']
    df_output['cp_slope_risk'] = df_output['cp_asymptomatic'] * (df_input['ST_Slope'] == 0).astype(int)
    df_output['cp_slope_interaction'] = df_input['ChestPainType'] * df_input['ST_Slope']
    df_output['cp_slope_ratio'] = df_input['ChestPainType'] / (df_input['ST_Slope'] + 1e-8)
    df_output['cp_slope_abs_diff'] = np.abs(df_input['ChestPainType'] - df_input['ST_Slope'])
    
    # --- Cholesterol ? ST_Slope interaction (strength 0.1145) ---
    df_output['chol_slope_product'] = df_input['Cholesterol'] * df_input['ST_Slope']
    df_output['chol_slope_ratio'] = df_input['Cholesterol'] / (df_input['ST_Slope'] + 1e-8)
    df_output['high_chol_downslope'] = df_output['chol_very_high'] * df_output['st_downsloping']
    df_output['high_chol_flat'] = df_output['chol_high'] * df_output['st_flat']
    df_output['chol_slope_normalized'] = df_input['Cholesterol'] / (df_input['ST_Slope'] * 100 + 1e-8)
    
    # --- Physiological ratios and products (non-redundant with originals) ---
    df_output['age_hr_ratio'] = df_input['Age'] / (df_input['MaxHR'] + 1e-8)
    df_output['bp_age_ratio'] = df_input['RestingBP'] / (df_input['Age'] + 1e-8)
    df_output['bp_chol_ratio'] = df_input['RestingBP'] / (df_input['Cholesterol'] + 1e-8)
    df_output['chol_bp_product'] = df_input['Cholesterol'] * df_input['RestingBP']
    df_output['chol_bp_ratio'] = df_input['Cholesterol'] / (df_input['RestingBP'] + 1e-8)
    df_output['age_chol_product'] = df_input['Age'] * df_input['Cholesterol']
    df_output['bp_chol_interaction'] = df_input['RestingBP'] * (df_input['Cholesterol'] > 200).astype(int)
    
    # --- Heart rate related (non-redundant) ---
    df_output['hr_reserve'] = (220 - df_input['Age']) - df_input['MaxHR']
    df_output['hr_reserve_low'] = (df_output['hr_reserve'] > 40).astype(int)
    df_output['hr_deficit'] = np.maximum(0, 190 - df_input['MaxHR'])
    df_output['hr_deficit_positive'] = (df_output['hr_deficit'] > 0).astype(int)
    df_output['hr_over_150'] = (df_input['MaxHR'] > 150).astype(int)
    df_output['maxhr_age_ratio'] = df_input['MaxHR'] / (df_input['Age'] + 1e-8)
    df_output['maxhr_chol_ratio'] = df_input['MaxHR'] / (df_input['Cholesterol'] + 1e-8)
    df_output['maxhr_oldpeak_ratio'] = df_input['MaxHR'] / (df_input['Oldpeak'] + 1e-8)
    df_output['chol_per_hr'] = df_input['Cholesterol'] / (df_input['MaxHR'] + 1e-8)
    df_output['bp_per_hr'] = df_input['RestingBP'] / (df_input['MaxHR'] + 1e-8)
    
    # --- Oldpeak interactions with slope ---
    df_output['oldpeak_slope_risk'] = df_input['Oldpeak'] * (df_input['ST_Slope'] == 0).astype(int)
    df_output['oldpeak_slope_combo'] = df_input['Oldpeak'] * df_input['ST_Slope']
    df_output['oldpeak_slope_ratio'] = df_input['Oldpeak'] / (df_input['ST_Slope'] + 1e
\end{lstlisting}

Three variants of this prompt occur in the runs. First, the role and instruction block alternates between two templates, as in \llmfe. Listing~\ref{lst:prompt} shows the domain template. In every third generation, including the initialisation, the operations template is used instead. It adds an \texttt{<Operators>} list of allowed row-wise operations (absolute value, logarithm, square root, square, sigmoid, rounding, reciprocal, the four arithmetic operations, minimum, maximum, products and ratios, numeric combination of two categorical columns, fixed-threshold indicators and \texttt{np.where} conditionals). \llmfe{} draws its operations template at random with probability $1/3$ per call, so both methods use it for about a third of their calls. Its list contains the cross-row operators \texttt{Frequency}, \texttt{GroupByThen*} and \texttt{Combine} that Sect.~\ref{sec:h2h} of the main text discusses, and ours replaces them with the row-wise operations above. Second, the initialisation prompt has no population summary and no report. Its parent block is titled ``Starting Template'' and holds the identity function, and its guidance is the single generic line ``Improve the selected solution by creating new domain-relevant features, or redesign with a completely new feature engineering strategy.'' Third, in the no-SHAP control the feature impact report is absent and the guidance block is that same generic line in every prompt, while the population summary, the parent's code and score and the failed-attempts block below remain.

The prompt in Listing~\ref{lst:prompt} has no failed-attempts block because nothing had failed on that fold by then. Listing~\ref{lst:failed} shows the block as it appears when a failure is on record, taken from the Claude-Haiku-5.5 run on \texttt{heart}, outer fold~2, prompt~62. The program generated by prompt~61 read a feature of the parent, \texttt{slope\_hr\_depr}, from \texttt{df\_input}, which holds only the original columns, and raised a \texttt{KeyError}. The block is inserted between the SHAP feedback report and the guidance. It shows the most recent failures of the fold, at most two, the code cut at 600 characters and the error at 250. The record is not cleared, so a failure stays visible until two newer failures replace it. Haiku failed rarely on this fold, and this block was repeated unchanged in every mutation prompt from 62 to 117, and from prompt 118 on a newer failure took its place.

\begin{lstlisting}[caption={The failed-attempts block of prompt~62 on \texttt{heart}, fold~2, with Claude-Haiku-5.5.},label={lst:failed}]
=== Recent Failed Attempts (do NOT repeat these specific mistakes) ===
Each block shows code that was just generated and the error it produced.
Diagnose the precise mistake and avoid it in your new function.

--- Attempt 1 ---
Code that failed:
```python
"""Creates row-wise ST_Slope interaction features with Oldpeak, Age, and slope_hr_depr, plus a non-redundant ChestPainType x exercise-angina synergy, using fixed numeric scales to capture ischemic risk combinations."""
df_output = df_input.copy()

# ST_Slope x Oldpeak: flat/downsloping ST with depression is a strong ischemia marker
df_output['slope_x_oldpeak'] = (2.0 - df_input['ST_Slope']) * df_input['Oldpeak']

# ST_Slope x Age: ischemic ST changes carry greater risk with advancing age
df_output['slope_x_age'] = (2.0 - df_input['ST_Slope']) * df_input['Age'] / 100.0

# ST_Slope x slope_hr_de
# ...<truncated>
```
Error: Execution error: 'slope_hr_depr'
\end{lstlisting}

\section{LLM Tokens and API Cost}
\label{sec:cost}
Table~\ref{tab:cost} gives the input and output tokens and the API cost of every arm over the whole benchmark, that is 22 datasets, five folds and 200 budget units per fold, or 21 datasets with GPT-3.5, where \texttt{communities} is excluded. For the two \shapevo{} arms the numbers are the tokens billed by the provider, which every call records. The \llmfe{} code records no usage, so its prompt is reconstructed as described for Table~\ref{tab:summary} of the main paper in the convergence section, counted with cl100k and divided by the ratio of the cl100k count to the billed count over all \shapevo{} calls of the same backbone, which is 1.00 for GPT-3.5, 0.93 for DeepSeek and 0.66 for Haiku input and 0.61 for Haiku output. The cost is the token total at the OpenRouter list price of the day of the runs. A \shapevo{} prompt is longer than an \llmfe{} prompt because it carries the SHAP report, by about 800 tokens with GPT-3.5 and 2,500 with DeepSeek, and its reply is shorter with GPT-3.5 and Haiku. One arm of the benchmark costs between 21 and 43 US dollars, and the three arms of a backbone are within a factor of 1.5 of each other. The sums over the table, 126 US dollars for GPT-3.5, 80 for DeepSeek and 83 for
Haiku, agree with the bills of Sect.~\ref{sec:harness} once the pilot and the aborted runs are added.

\begin{table}[h]
\centering
\small
\caption{Input and output tokens and API cost per method and backbone over the whole benchmark (all datasets and folds, 200 budget units per fold). For the \shapevo{} arms the tokens are those billed by the provider. For \llmfe{} the reconstructed prompt and the stored reply are counted with cl100k and calibrated to the billed count of the same backbone. Cost at the OpenRouter list price of the day of the runs, in US dollars per million input and output tokens 0.5 and 1.5 for GPT-3.5-Turbo, 0.018 and 0.64 for DeepSeek-V4-Flash, 0.1 and 0.5 for Claude-Haiku-5.5.}
\label{tab:cost}
\begin{tabular}{llrrrrr}
\toprule
Backbone & Method & \multicolumn{2}{c}{Tokens per call} & \multicolumn{2}{c}{Total (M)} & Cost \\
\cmidrule(lr){3-4}\cmidrule(lr){5-6}
 & & in & out & in & out & (USD) \\
\midrule
GPT-3.5-Turbo & \llmfe & 2,856 & 356 & 60.6 & 7.6 & 41.62 \\
 & \shapevo & 3,639 & 170 & 76.0 & 3.6 & 43.35 \\
 & no-SHAP control & 3,336 & 190 & 69.7 & 4.0 & 40.81 \\
DeepSeek-V4-Flash & \llmfe & 5,541 & 1,297 & 123.1 & 28.8 & 20.67 \\
 & \shapevo & 8,026 & 1,852 & 175.7 & 40.5 & 29.10 \\
 & no-SHAP control & 6,611 & 2,006 & 144.7 & 43.9 & 30.71 \\
Claude-Haiku-5.5 & \llmfe & 6,609 & 1,341 & 146.8 & 29.8 & 29.58 \\
 & \shapevo & 6,933 & 817 & 151.8 & 17.9 & 24.12 \\
 & no-SHAP control & 6,729 & 1,351 & 147.3 & 29.6 & 29.52 \\
\bottomrule
\end{tabular}

\end{table}

\section{Held-Out Head-to-Head Scores}
\label{sec:h2h_supp}
Table~\ref{tab:h2h} gives the per-dataset held-out scores behind the wins, ties and losses of Table~\ref{tab:summary} of the main text, discussed in Sect.~\ref{sec:h2h} there.

\begin{table}[h]
\centering
\caption{Held-out head-to-head results with 5-fold outer CV (seed 42), 200 evaluations per fold and the top-3 ensemble scored by \llmfe's official evaluator, mean $\pm$ std over folds. $^{R}$ marks regression, reported as N-RMSE (lower is better). Bold marks the winner when $|\Delta|>0.01$, where $\Delta>0$ favours \shapevo. The 
rule is applied to the unrounded scores, so a displayed $\pm0.010$ can fall on either side of it: GPT-3.5 \texttt{housing} ($+0.0104$) and DeepSeek \texttt{tic-tac-toe} ($+0.0104$) count as wins, GPT-3.5 
\texttt{eucalyptus} ($+0.0095$) and \texttt{crab} ($-0.0096$) as ties. With GPT-3.5 the \llmfe{} mean of \texttt{communities} is over one fold (context overflow, Sect.~\ref{sec:h2h} of the main text). With DeepSeek and Haiku the \texttt{car} and \texttt{bike} means of \llmfe, and with Haiku its \texttt{tic-tac-toe} mean, include folds
on which its ensemble collapsed.}
\label{tab:h2h}
\scriptsize
\resizebox{\textwidth}{!}{\begin{tabular}{lrrrrrrrrr}
\toprule
Dataset & \multicolumn{3}{c}{GPT-3.5-Turbo} & \multicolumn{3}{c}{DeepSeek-V4-Flash} & \multicolumn{3}{c}{Claude-Haiku-5.5} \\
 & LLM-FE & SHAPEvo-FE & $\Delta$ & LLM-FE & SHAPEvo-FE & $\Delta$ & LLM-FE & SHAPEvo-FE & $\Delta$ \\
\midrule
adult & 0.872$\pm$0.004 & 0.872$\pm$0.002 & +0.000 & 0.868$\pm$0.010 & 0.873$\pm$0.002 & +0.005 & 0.874$\pm$0.003 & 0.874$\pm$0.003 & +0.000 \\
balance-scale & 1.000$\pm$0.000 & 1.000$\pm$0.000 & +0.000 & 1.000$\pm$0.000 & 1.000$\pm$0.000 & +0.000 & 1.000$\pm$0.000 & 1.000$\pm$0.000 & +0.000 \\
bank & 0.907$\pm$0.003 & 0.906$\pm$0.003 & -0.001 & 0.907$\pm$0.004 & 0.907$\pm$0.002 & -0.000 & 0.894$\pm$0.019 & \textbf{0.909$\pm$0.005} & \textbf{+0.015} \\
blood & 0.741$\pm$0.030 & 0.735$\pm$0.029 & -0.005 & 0.749$\pm$0.030 & 0.755$\pm$0.026 & +0.007 & 0.751$\pm$0.020 & 0.759$\pm$0.033 & +0.008 \\
breast-w & 0.957$\pm$0.010 & 0.959$\pm$0.003 & +0.001 & 0.971$\pm$0.011 & 0.967$\pm$0.004 & -0.004 & 0.966$\pm$0.009 & 0.961$\pm$0.012 & -0.004 \\
car & 1.000$\pm$0.000 & 0.999$\pm$0.003 & -0.001 & 0.908$\pm$0.174 & \textbf{1.000$\pm$0.000} & \textbf{+0.092} & 0.957$\pm$0.093 & \textbf{0.998$\pm$0.004} & \textbf{+0.041} \\
cmc & 0.515$\pm$0.015 & 0.521$\pm$0.010 & +0.006 & \textbf{0.526$\pm$0.027} & 0.511$\pm$0.017 & \textbf{-0.015} & \textbf{0.521$\pm$0.022} & 0.495$\pm$0.021 & \textbf{-0.027} \\
communities & 0.709$\pm$0.000 & 0.707$\pm$0.017 & -0.003 & \textbf{0.706$\pm$0.016} & 0.694$\pm$0.015 & \textbf{-0.012} & 0.706$\pm$0.022 & 0.708$\pm$0.022 & +0.002 \\
credit-g & 0.758$\pm$0.017 & 0.749$\pm$0.018 & -0.009 & \textbf{0.758$\pm$0.019} & 0.740$\pm$0.010 & \textbf{-0.018} & 0.741$\pm$0.021 & 0.740$\pm$0.017 & -0.001 \\
eucalyptus & 0.663$\pm$0.028 & 0.673$\pm$0.029 & +0.010 & 0.667$\pm$0.034 & 0.674$\pm$0.018 & +0.007 & 0.671$\pm$0.039 & 0.679$\pm$0.021 & +0.008 \\
heart & 0.862$\pm$0.020 & 0.857$\pm$0.023 & -0.004 & 0.852$\pm$0.024 & 0.854$\pm$0.031 & +0.002 & 0.855$\pm$0.023 & 0.861$\pm$0.026 & +0.005 \\
junglechess & \textbf{0.980$\pm$0.004} & 0.954$\pm$0.026 & \textbf{-0.026} & 0.997$\pm$0.001 & 0.997$\pm$0.001 & +0.000 & 0.992$\pm$0.002 & 0.993$\pm$0.001 & +0.001 \\
myocardial & \textbf{0.790$\pm$0.027} & 0.776$\pm$0.031 & \textbf{-0.015} & 0.796$\pm$0.019 & 0.789$\pm$0.015 & -0.007 & 0.780$\pm$0.033 & 0.778$\pm$0.026 & -0.001 \\
pc1 & 0.940$\pm$0.017 & 0.931$\pm$0.005 & -0.009 & 0.931$\pm$0.002 & 0.938$\pm$0.010 & +0.006 & 0.932$\pm$0.013 & 0.931$\pm$0.017 & -0.002 \\
synthetic & 0.872$\pm$0.003 & 0.865$\pm$0.007 & -0.007 & 0.867$\pm$0.007 & 0.871$\pm$0.005 & +0.004 & 0.868$\pm$0.008 & 0.871$\pm$0.004 & +0.003 \\
tic-tac-toe & 0.998$\pm$0.005 & 0.990$\pm$0.010 & -0.008 & 0.987$\pm$0.011 & \textbf{0.998$\pm$0.005} & \textbf{+0.010} & 0.903$\pm$0.194 & \textbf{0.999$\pm$0.002} & \textbf{+0.096} \\
vehicle & 0.759$\pm$0.021 & 0.766$\pm$0.030 & +0.007 & 0.775$\pm$0.018 & \textbf{0.801$\pm$0.033} & \textbf{+0.026} & 0.772$\pm$0.026 & \textbf{0.790$\pm$0.017} & \textbf{+0.018} \\
bike$^{R}$ & 0.224$\pm$0.037 & \textbf{0.211$\pm$0.006} & \textbf{+0.013} & 0.297$\pm$0.116 & \textbf{0.204$\pm$0.004} & \textbf{+0.094} & 0.244$\pm$0.061 & \textbf{0.202$\pm$0.006} & \textbf{+0.042} \\
crab$^{R}$ & 0.222$\pm$0.014 & 0.232$\pm$0.015 & -0.010 & 0.218$\pm$0.015 & 0.224$\pm$0.013 & -0.006 & 0.220$\pm$0.013 & 0.220$\pm$0.015 & +0.000 \\
housing$^{R}$ & 0.228$\pm$0.030 & \textbf{0.218$\pm$0.008} & \textbf{+0.010} & 0.245$\pm$0.017 & \textbf{0.211$\pm$0.005} & \textbf{+0.034} & 0.235$\pm$0.024 & \textbf{0.214$\pm$0.011} & \textbf{+0.021} \\
insurance$^{R}$ & 0.384$\pm$0.028 & 0.385$\pm$0.023 & -0.001 & 0.396$\pm$0.047 & \textbf{0.380$\pm$0.034} & \textbf{+0.016} & 0.387$\pm$0.030 & 0.381$\pm$0.021 & +0.006 \\
wine$^{R}$ & 0.105$\pm$0.001 & 0.108$\pm$0.001 & -0.003 & 0.105$\pm$0.001 & 0.106$\pm$0.003 & -0.001 & 0.105$\pm$0.002 & 0.106$\pm$0.002 & -0.001 \\
\midrule
SHAPEvo-FE wins / ties / losses & \multicolumn{3}{c}{W 2 / T 18 / L 2} & \multicolumn{3}{c}{W 6 / T 13 / L 3} & \multicolumn{3}{c}{W 6 / T 15 / L 1} \\
\bottomrule
\end{tabular}}
\end{table}

Which programs form the Haiku \llmfe{} ensemble on the \texttt{tic-tac-toe} and \texttt{car} folds that collapse is partly a matter of chance. Several programs tie for the best inner-CV score on these folds and the official scorer breaks ties by file order. Choosing the earliest generated programs among the tied ones instead avoids both collapses and turns these two wins into ties. We keep the official behaviour, since every \llmfe{} number in this paper comes from it.

\end{document}